%% file: main.tex
\documentclass{IEEEtran}
\ifCLASSOPTIONcompsoc
  \usepackage[nocompress]{cite}
\else
  \usepackage{cite}
\fi
\ifCLASSINFOpdf
  \usepackage[pdftex]{graphicx}
\else
\fi
\usepackage{amsmath}
\usepackage{float}
\usepackage{url}

\usepackage{multirow}
\usepackage{array}

\usepackage[table]{xcolor}

\newcommand{\ceff}{$C_\text{eff}$}
\newcommand{\ctotal}{$C_\text{total}$}
\newcommand{\slew}{$t_\Delta$}
\newcommand{\rd}{$R_d$}
\newcommand{\wireres}{$R_w$}
\newcommand{\wirecap}{$C_w$}
\newcommand{\pincap}{$C_p$}
\newcommand{\driverflag}{$f_d$}
\newcommand{\fanoutflag}{$f_f$}
\newcommand{\upres}{$R_u$}
\newcommand{\downcap}{$C_d$}
\newcommand{\numhops}{$h_d$}
\newcommand{\vl}{$V_\text{L}$}
\newcommand{\vh}{$V_\text{H}$}
\newcommand{\pimodel}{$\pi$-model}
\newcommand{\rcnetwork}{$G_\text{RC}$}
\newcommand{\gnngraph}{$G_\text{GNN}$}
\newcommand{\nodenode}{$\mathcal{V}_n$}
\newcommand{\edgenode}{$\mathcal{V}_e$}
\newcommand{\nodenodepin}{$\mathcal{V}_{np}$}
\newcommand{\nodenodejunc}{$\mathcal{V}_{nj}$}
\newcommand{\openroadtypical}{$\text{OpenROAD}_\text{typical}$}
\newcommand{\openroadfast}{$\text{OpenROAD}_\text{fast}$}
\newcommand{\openroadslow}{$\text{OpenROAD}_\text{slow}$}

\begin{document}
%
\title{Effective Capacitance Modeling\\Using Graph Neural Networks}

\author{\IEEEauthorblockN{Eren Dogan,
Matthew R. Guthaus }\\
\IEEEauthorblockA{Computer Science and Engineering\\ University of California Santa Cruz, Santa Cruz, CA 95064\\ 
\{erdogan, mrg\}@ucsc.edu}
\vspace{-0.7cm}
}


%


\newcommand{\todo}[1]{{\bf TODO: {#1}}\\}
\newcommand*\rot{\rotatebox{90}}
\input{variables}

\maketitle

\begin{abstract}
Static timing analysis is a crucial stage in the VLSI design flow that verifies the timing correctness of circuits. Timing analysis depends on the placement and routing of the design, but at the same time, placement and routing efficiency depend on the final timing performance. VLSI design flows can benefit from timing-related prediction to better perform the earlier stages of the design flow. Effective capacitance is an essential input for gate delay calculation, and finding exact values requires routing or routing estimates. In this work, we propose the first GNN-based post-layout effective capacitance modeling method, GNN-\ceff{}, that achieves significant speed gains due to GPU parallelization while also providing better accuracy than current heuristics. GNN-\ceff{} parallelization achieves \OpenroadTypicalSpeedupOpenSerialDartu{} speedup on real-life benchmarks over the state-of-the-art method run serially.
\end{abstract}


%
\IEEEpeerreviewmaketitle

\input{intro}
\input{background}
\input{proposed}

\input{results}
\input{conclusion}







%
\clearpage
{\small
\bibliographystyle{ieeetr}
\bibliography{main}
}

\end{document}

%% file: variables.tex
\def \modelConvLayer {GAT}
\def \modelConvLayerRef {\cite{gat}}
\def \modelNumConvLayers {3}
\def \modelNumConvChannels {32}
\def \modelNumAttentionHeads {12}
\def \modelConvActivationFunc {ELU}
\def \modelNumLinearLayers {3}
\def \modelNumLinearChannels {64}
\def \modelLinearActivationFunc {ELU}
\def \modelLinearActivationFuncFinal {Sigmoid}
\def \modelLayerDropout {0.0001}
\def \modelAttentionDropout {0.0001}
\def \trainBatchSize {100}
\def \trainLearningRate {0.001}
\def \trainEpochs {300}
\def \trainPatience {10}
\def \trainLRPatience {5}
\def \trainDelta {1e-10}

\def \SyntheticFailratio{6.19}                                
\def \SyntheticAllcasesAbserrorGnnceffMean {0.24}
\def \SyntheticAllcasesAbserrorGnnceffMax {14.84}
\def \SyntheticAllcasesAbserrorGnnceffStdev {0.36}
\def \SyntheticAllcasesAbserrorDartuMean {7.85}
\def \SyntheticAllcasesAbserrorDartuMax {191.27}
\def \SyntheticAllcasesAbserrorDartuStdev {23.26}
\def \SyntheticFailAbserrorGnnceffMean {0.60}
\def \SyntheticFailAbserrorGnnceffMax {13.41}
\def \SyntheticFailAbserrorGnnceffStdev {0.64}
\def \SyntheticFailAbserrorDartuMean {92.42}
\def \SyntheticFailAbserrorDartuMax {191.27}
\def \SyntheticFailAbserrorDartuStdev {31.98}
\def \SyntheticNofailAbserrorGnnceffMean {0.22}
\def \SyntheticNofailAbserrorGnnceffMax {14.84}
\def \SyntheticNofailAbserrorGnnceffStdev {0.31}
\def \SyntheticNofailAbserrorDartuMean {2.27}
\def \SyntheticNofailAbserrorDartuMax {39.19}
\def \SyntheticNofailAbserrorDartuStdev {2.53}
\def \SyntheticAllcasesAbserrorratioGnnceffMean {0.51}
\def \SyntheticAllcasesAbserrorratioGnnceffMax {419.28}
\def \SyntheticAllcasesAbserrorratioGnnceffStdev {1.66}
\def \SyntheticAllcasesAbserrorratioDartuMean {27.54}
\def \SyntheticAllcasesAbserrorratioDartuMax {25,558.53}
\def \SyntheticAllcasesAbserrorratioDartuStdev {154.40}
\def \SyntheticFailAbserrorratioGnnceffMean {2.63}
\def \SyntheticFailAbserrorratioGnnceffMax {419.28}
\def \SyntheticFailAbserrorratioGnnceffStdev {5.92}
\def \SyntheticFailAbserrorratioDartuMean {389.19}
\def \SyntheticFailAbserrorratioDartuMax {25,558.53}
\def \SyntheticFailAbserrorratioDartuStdev {495.40}
\def \SyntheticNofailAbserrorratioGnnceffMean {0.37}
\def \SyntheticNofailAbserrorratioGnnceffMax {34.19}
\def \SyntheticNofailAbserrorratioGnnceffStdev {0.55}
\def \SyntheticNofailAbserrorratioDartuMean {3.68}
\def \SyntheticNofailAbserrorratioDartuMax {150.96}
\def \SyntheticNofailAbserrorratioDartuStdev {4.59}
\def \SyntheticAllcasesDelayabserrorGnnceffMean {1.16}
\def \SyntheticAllcasesDelayabserrorGnnceffMax {41.25}
\def \SyntheticAllcasesDelayabserrorGnnceffStdev {1.63}
\def \SyntheticAllcasesDelayabserrorDartuMean {24.69} 
\def \SyntheticAllcasesDelayabserrorDartuMax {1,448.52}
\def \SyntheticAllcasesDelayabserrorDartuStdev {62.91}
\def \SyntheticFailDelayabserrorGnnceffMean {1.33}       
\def \SyntheticFailDelayabserrorGnnceffMax {27.90}       
\def \SyntheticFailDelayabserrorGnnceffStdev {1.84} 
\def \SyntheticFailDelayabserrorDartuMean {203.87}            
\def \SyntheticFailDelayabserrorDartuMax {1,448.52}
\def \SyntheticFailDelayabserrorDartuStdev {163.14}
\def \SyntheticNofailDelayabserrorGnnceffMean {1.15}
\def \SyntheticNofailDelayabserrorGnnceffMax {41.25}
\def \SyntheticNofailDelayabserrorGnnceffStdev {1.61}
\def \SyntheticNofailDelayabserrorDartuMean {12.87}
\def \SyntheticNofailDelayabserrorDartuMax {196.91}
\def \SyntheticNofailDelayabserrorDartuStdev {14.30}
\def \SyntheticAllcasesTimeGnnceffMean {75}
\def \SyntheticAllcasesTimeGnnceffMax {76}
\def \SyntheticAllcasesTimeGnnceffStdev {1}
\def \SyntheticFailTimeGnnceffMean {75}
\def \SyntheticFailTimeGnnceffMax {76}
\def \SyntheticFailTimeGnnceffStdev {1}
\def \SyntheticNofailTimeGnnceffMean {75}
\def \SyntheticNofailTimeGnnceffMax {76}
\def \SyntheticNofailTimeGnnceffStdev {1}
\def \SyntheticAllcasesTimeSyntSerialDartuMean {15,751}
\def \SyntheticAllcasesTimeSyntSerialDartuMax {1,032,252}
\def \SyntheticAllcasesTimeSyntSerialDartuStdev {9,310}
\def \SyntheticFailTimeSyntSerialDartuMean {9,581}
\def \SyntheticFailTimeSyntSerialDartuMax {678,120}
\def \SyntheticFailTimeSyntSerialDartuStdev {7,062}
\def \SyntheticNofailTimeSyntSerialDartuMean {16,159}
\def \SyntheticNofailTimeSyntSerialDartuMax {1,032,252}
\def \SyntheticNofailTimeSyntSerialDartuStdev {9,297}
\def \SyntheticAllcasesTimeOpenSerialDartuMean {131,262}
\def \SyntheticAllcasesTimeOpenSerialDartuMax {8,602,098}
\def \SyntheticAllcasesTimeOpenSerialDartuStdev {77,585}
\def \SyntheticFailTimeOpenSerialDartuMean {79,843}
\def \SyntheticFailTimeOpenSerialDartuMax {5,651,000}
\def \SyntheticFailTimeOpenSerialDartuStdev {58,846}
\def \SyntheticNofailTimeOpenSerialDartuMean {134,655}
\def \SyntheticNofailTimeOpenSerialDartuMax {8,602,098}
\def \SyntheticNofailTimeOpenSerialDartuStdev {77,473}
\def \SyntheticAllcasesTimeSyntParallelDartuMean {246}
\def \SyntheticAllcasesTimeSyntParallelDartuMax {16,129}
\def \SyntheticAllcasesTimeSyntParallelDartuStdev {145}
\def \SyntheticFailTimeSyntParallelDartuMean {150}
\def \SyntheticFailTimeSyntParallelDartuMax {10,596}
\def \SyntheticFailTimeSyntParallelDartuStdev {110}
\def \SyntheticNofailTimeSyntParallelDartuMean {252}
\def \SyntheticNofailTimeSyntParallelDartuMax {16,129}
\def \SyntheticNofailTimeSyntParallelDartuStdev {145}
\def \SyntheticAllcasesTimeOpenParallelDartuMean {2,051}
\def \SyntheticAllcasesTimeOpenParallelDartuMax {134,408}
\def \SyntheticAllcasesTimeOpenParallelDartuStdev {1,212}
\def \SyntheticFailTimeOpenParallelDartuMean {1,248}
\def \SyntheticFailTimeOpenParallelDartuMax {88,297}
\def \SyntheticFailTimeOpenParallelDartuStdev {919}
\def \SyntheticNofailTimeOpenParallelDartuMean {2,104}
\def \SyntheticNofailTimeOpenParallelDartuMax {134,408}
\def \SyntheticNofailTimeOpenParallelDartuStdev {1,211}
\def \SyntheticAllcasesTimeSyntSerialNgspiceMean {64,606,507}
\def \SyntheticAllcasesTimeSyntSerialNgspiceMax {188,279,065}
\def \SyntheticAllcasesTimeSyntSerialNgspiceStdev {22,730,152} 
\def \SyntheticFailTimeSyntSerialNgspiceMean {73,450,869}
\def \SyntheticFailTimeSyntSerialNgspiceMax {188,279,065}
\def \SyntheticFailTimeSyntSerialNgspiceStdev {21,608,934}
\def \SyntheticNofailTimeSyntSerialNgspiceMean {64,022,959}
\def \SyntheticNofailTimeSyntSerialNgspiceMax {179,491,907}
\def \SyntheticNofailTimeSyntSerialNgspiceStdev {22,681,257}
\def \SyntheticAllcasesTimeOpenSerialNgspiceMean {538,387,560} 
\def \SyntheticAllcasesTimeOpenSerialNgspiceMax {1,568,992,208}
\def \SyntheticAllcasesTimeOpenSerialNgspiceStdev {189,417,933}
\def \SyntheticFailTimeOpenSerialNgspiceMean {612,090,579}
\def \SyntheticFailTimeOpenSerialNgspiceMax {1,568,992,208}
\def \SyntheticFailTimeOpenSerialNgspiceStdev {180,074,448}
\def \SyntheticNofailTimeOpenSerialNgspiceMean {533,524,656}
\def \SyntheticNofailTimeOpenSerialNgspiceMax {1,495,765,896}
\def \SyntheticNofailTimeOpenSerialNgspiceStdev {189,010,474}
\def \SyntheticAllcasesTimeSyntParallelNgspiceMean {1,009,477} 
\def \SyntheticAllcasesTimeSyntParallelNgspiceMax {2,941,860}
\def \SyntheticAllcasesTimeSyntParallelNgspiceStdev {355,159}
\def \SyntheticFailTimeSyntParallelNgspiceMean {1,147,670}
\def \SyntheticFailTimeSyntParallelNgspiceMax {2,941,860}
\def \SyntheticFailTimeSyntParallelNgspiceStdev {337,640}
\def \SyntheticNofailTimeSyntParallelNgspiceMean {1,000,359}
\def \SyntheticNofailTimeSyntParallelNgspiceMax {2,804,561}
\def \SyntheticNofailTimeSyntParallelNgspiceStdev {354,395}
\def \SyntheticAllcasesTimeOpenParallelNgspiceMean {8,412,306}
\def \SyntheticAllcasesTimeOpenParallelNgspiceMax {24,515,503}
\def \SyntheticAllcasesTimeOpenParallelNgspiceStdev {2,959,655}
\def \SyntheticFailTimeOpenParallelNgspiceMean {9,563,915}
\def \SyntheticFailTimeOpenParallelNgspiceMax {24,515,503}
\def \SyntheticFailTimeOpenParallelNgspiceStdev {2,813,663}
\def \SyntheticNofailTimeOpenParallelNgspiceMean {8,336,323}
\def \SyntheticNofailTimeOpenParallelNgspiceMax {23,371,342}
\def \SyntheticNofailTimeOpenParallelNgspiceStdev {2,953,289}
\def \SyntheticSpeedupOpenSerialDartu {1,801x}
\def \SyntheticSpeedupOpenParallelDartu {28x}
\def \SyntheticSpeedupSyntSerialDartu {216x}
\def \SyntheticSpeedupSyntParallelDartu {3x}
\def \SyntheticSpeedupOpenSerialNgspice {7,137,693x}
\def \SyntheticSpeedupOpenParallelNgspice {111,526x}
\def \SyntheticSpeedupSyntSerialNgspice {856,523x}
\def \SyntheticSpeedupSyntParallelNgspice {13,383x}

\def \OpenroadTypicalFailratio{0.04}                                
\def \OpenroadTypicalAllcasesAbserrorGnnceffMean {0.01}    
\def \OpenroadTypicalAllcasesAbserrorGnnceffMax {5.28}         
\def \OpenroadTypicalAllcasesAbserrorGnnceffStdev {0.06}      
\def \OpenroadTypicalAllcasesAbserrorDartuMean {0.01}          
\def \OpenroadTypicalAllcasesAbserrorDartuMax {21.15}      
\def \OpenroadTypicalAllcasesAbserrorDartuStdev {0.24}    
\def \OpenroadTypicalFailAbserrorGnnceffMean {0.92}        
\def \OpenroadTypicalFailAbserrorGnnceffMax {4.99}           
\def \OpenroadTypicalFailAbserrorGnnceffStdev {1.11}        
\def \OpenroadTypicalFailAbserrorDartuMean {10.24}           
\def \OpenroadTypicalFailAbserrorDartuMax {21.15}               
\def \OpenroadTypicalFailAbserrorDartuStdev {4.83}             
\def \OpenroadTypicalNofailAbserrorGnnceffMean {0.01}           
\def \OpenroadTypicalNofailAbserrorGnnceffMax {5.28}        
\def \OpenroadTypicalNofailAbserrorGnnceffStdev {0.06}      
\def \OpenroadTypicalNofailAbserrorDartuMean {0.01}         
\def \OpenroadTypicalNofailAbserrorDartuMax {10.54}           
\def \OpenroadTypicalNofailAbserrorDartuStdev {0.08}         
\def \OpenroadTypicalAllcasesAbserrorratioGnnceffMean {0.22}  
\def \OpenroadTypicalAllcasesAbserrorratioGnnceffMax {116.25} 
\def \OpenroadTypicalAllcasesAbserrorratioGnnceffStdev {1.12}  
\def \OpenroadTypicalAllcasesAbserrorratioDartuMean {0.39}     
\def \OpenroadTypicalAllcasesAbserrorratioDartuMax {410.33}
\def \OpenroadTypicalAllcasesAbserrorratioDartuStdev {4.51}  
\def \OpenroadTypicalFailAbserrorratioGnnceffMean {19.06}    
\def \OpenroadTypicalFailAbserrorratioGnnceffMax {116.25}     
\def \OpenroadTypicalFailAbserrorratioGnnceffStdev {24.09}   
\def \OpenroadTypicalFailAbserrorratioDartuMean {194.51}     
\def \OpenroadTypicalFailAbserrorratioDartuMax {410.33}         
\def \OpenroadTypicalFailAbserrorratioDartuStdev {91.05}        
\def \OpenroadTypicalNofailAbserrorratioGnnceffMean {0.22}       
\def \OpenroadTypicalNofailAbserrorratioGnnceffMax {65.20}  
\def \OpenroadTypicalNofailAbserrorratioGnnceffStdev {0.94}
\def \OpenroadTypicalNofailAbserrorratioDartuMean {0.32}   
\def \OpenroadTypicalNofailAbserrorratioDartuMax {273.04}     
\def \OpenroadTypicalNofailAbserrorratioDartuStdev {1.77}     
\def \OpenroadTypicalAllcasesDelayabserrorGnnceffMean {0.01}   
\def \OpenroadTypicalAllcasesDelayabserrorGnnceffMax {9.11}
\def \OpenroadTypicalAllcasesDelayabserrorGnnceffStdev {0.09}
\def \OpenroadTypicalAllcasesDelayabserrorDartuMean {0.02}
\def \OpenroadTypicalAllcasesDelayabserrorDartuMax {37.95}
\def \OpenroadTypicalAllcasesDelayabserrorDartuStdev {0.29} 
\def \OpenroadTypicalFailDelayabserrorGnnceffMean {1.18}     
\def \OpenroadTypicalFailDelayabserrorGnnceffMax {9.11}      
\def \OpenroadTypicalFailDelayabserrorGnnceffStdev {1.89}
\def \OpenroadTypicalFailDelayabserrorDartuMean {10.93}  
\def \OpenroadTypicalFailDelayabserrorDartuMax {37.95}
\def \OpenroadTypicalFailDelayabserrorDartuStdev {7.84}
\def \OpenroadTypicalNofailDelayabserrorGnnceffMean {0.01}
\def \OpenroadTypicalNofailDelayabserrorGnnceffMax {6.03}
\def \OpenroadTypicalNofailDelayabserrorGnnceffStdev {0.08}
\def \OpenroadTypicalNofailDelayabserrorDartuMean {0.02}
\def \OpenroadTypicalNofailDelayabserrorDartuMax {10.05}
\def \OpenroadTypicalNofailDelayabserrorDartuStdev {0.10}
\def \OpenroadTypicalAllcasesTimeGnnceffMean {49}
\def \OpenroadTypicalAllcasesTimeGnnceffMax {49}
\def \OpenroadTypicalAllcasesTimeGnnceffStdev {0}
\def \OpenroadTypicalFailTimeGnnceffMean {49}
\def \OpenroadTypicalFailTimeGnnceffMax {49}
\def \OpenroadTypicalFailTimeGnnceffStdev {0}
\def \OpenroadTypicalNofailTimeGnnceffMean {49}
\def \OpenroadTypicalNofailTimeGnnceffMax {49}
\def \OpenroadTypicalNofailTimeGnnceffStdev {0}
\def \OpenroadTypicalAllcasesTimeSyntSerialDartuMean {5,492}
\def \OpenroadTypicalAllcasesTimeSyntSerialDartuMax {47,785}
\def \OpenroadTypicalAllcasesTimeSyntSerialDartuStdev {1,696}
\def \OpenroadTypicalFailTimeSyntSerialDartuMean {6,118}
\def \OpenroadTypicalFailTimeSyntSerialDartuMax {14,620}
\def \OpenroadTypicalFailTimeSyntSerialDartuStdev {1,852}
\def \OpenroadTypicalNofailTimeSyntSerialDartuMean {5,492}
\def \OpenroadTypicalNofailTimeSyntSerialDartuMax {47,785}
\def \OpenroadTypicalNofailTimeSyntSerialDartuStdev {1,696}
\def \OpenroadTypicalAllcasesTimeOpenSerialDartuMean {45,767}
\def \OpenroadTypicalAllcasesTimeOpenSerialDartuMax {398,208}
\def \OpenroadTypicalAllcasesTimeOpenSerialDartuStdev {14,136}
\def \OpenroadTypicalFailTimeOpenSerialDartuMean {50,987}
\def \OpenroadTypicalFailTimeOpenSerialDartuMax {121,833}
\def \OpenroadTypicalFailTimeOpenSerialDartuStdev {15,434}
\def \OpenroadTypicalNofailTimeOpenSerialDartuMean {45,765}
\def \OpenroadTypicalNofailTimeOpenSerialDartuMax {398,208}
\def \OpenroadTypicalNofailTimeOpenSerialDartuStdev {14,136}
\def \OpenroadTypicalAllcasesTimeSyntParallelDartuMean {86}
\def \OpenroadTypicalAllcasesTimeSyntParallelDartuMax {747}
\def \OpenroadTypicalAllcasesTimeSyntParallelDartuStdev {27}
\def \OpenroadTypicalFailTimeSyntParallelDartuMean {96}
\def \OpenroadTypicalFailTimeSyntParallelDartuMax {228}
\def \OpenroadTypicalFailTimeSyntParallelDartuStdev {29}
\def \OpenroadTypicalNofailTimeSyntParallelDartuMean {86}
\def \OpenroadTypicalNofailTimeSyntParallelDartuMax {747}
\def \OpenroadTypicalNofailTimeSyntParallelDartuStdev {27}
\def \OpenroadTypicalAllcasesTimeOpenParallelDartuMean {715}
\def \OpenroadTypicalAllcasesTimeOpenParallelDartuMax {6,222}
\def \OpenroadTypicalAllcasesTimeOpenParallelDartuStdev {221}
\def \OpenroadTypicalFailTimeOpenParallelDartuMean {797}
\def \OpenroadTypicalFailTimeOpenParallelDartuMax {1,904}
\def \OpenroadTypicalFailTimeOpenParallelDartuStdev {241}
\def \OpenroadTypicalNofailTimeOpenParallelDartuMean {715}
\def \OpenroadTypicalNofailTimeOpenParallelDartuMax {6,222}
\def \OpenroadTypicalNofailTimeOpenParallelDartuStdev {221}
\def \OpenroadTypicalAllcasesTimeSyntSerialNgspiceMean {51,420}
\def \OpenroadTypicalAllcasesTimeSyntSerialNgspiceMax {91,156}
\def \OpenroadTypicalAllcasesTimeSyntSerialNgspiceStdev {7,714}
\def \OpenroadTypicalFailTimeSyntSerialNgspiceMean {51,729}
\def \OpenroadTypicalFailTimeSyntSerialNgspiceMax {64,386}
\def \OpenroadTypicalFailTimeSyntSerialNgspiceStdev {7,516}
\def \OpenroadTypicalNofailTimeSyntSerialNgspiceMean {51,419}
\def \OpenroadTypicalNofailTimeSyntSerialNgspiceMax {91,156}
\def \OpenroadTypicalNofailTimeSyntSerialNgspiceStdev {7,714}
\def \OpenroadTypicalAllcasesTimeOpenSerialNgspiceMean {428,497}
\def \OpenroadTypicalAllcasesTimeOpenSerialNgspiceMax {759,630}
\def \OpenroadTypicalAllcasesTimeOpenSerialNgspiceStdev {64,280}
\def \OpenroadTypicalFailTimeOpenSerialNgspiceMean {431,074}
\def \OpenroadTypicalFailTimeOpenSerialNgspiceMax {536,551}
\def \OpenroadTypicalFailTimeOpenSerialNgspiceStdev {62,630}
\def \OpenroadTypicalNofailTimeOpenSerialNgspiceMean {428,494}
\def \OpenroadTypicalNofailTimeOpenSerialNgspiceMax {759,630}
\def \OpenroadTypicalNofailTimeOpenSerialNgspiceStdev {64,282}
\def \OpenroadTypicalAllcasesTimeSyntParallelNgspiceMean {803}
\def \OpenroadTypicalAllcasesTimeSyntParallelNgspiceMax {1,424}
\def \OpenroadTypicalAllcasesTimeSyntParallelNgspiceStdev {121}
\def \OpenroadTypicalFailTimeSyntParallelNgspiceMean {808}
\def \OpenroadTypicalFailTimeSyntParallelNgspiceMax {1,006}
\def \OpenroadTypicalFailTimeSyntParallelNgspiceStdev {117}
\def \OpenroadTypicalNofailTimeSyntParallelNgspiceMean {803}
\def \OpenroadTypicalNofailTimeSyntParallelNgspiceMax {1,424}
\def \OpenroadTypicalNofailTimeSyntParallelNgspiceStdev {121}
\def \OpenroadTypicalAllcasesTimeOpenParallelNgspiceMean {6,695}
\def \OpenroadTypicalAllcasesTimeOpenParallelNgspiceMax {11,869}
\def \OpenroadTypicalAllcasesTimeOpenParallelNgspiceStdev {1,004}
\def \OpenroadTypicalFailTimeOpenParallelNgspiceMean {6,736}
\def \OpenroadTypicalFailTimeOpenParallelNgspiceMax {8,384}
\def \OpenroadTypicalFailTimeOpenParallelNgspiceStdev {979}
\def \OpenroadTypicalNofailTimeOpenParallelNgspiceMean {6,695}
\def \OpenroadTypicalNofailTimeOpenParallelNgspiceMax {11,869}
\def \OpenroadTypicalNofailTimeOpenParallelNgspiceStdev {1,004}
\def \OpenroadTypicalSpeedupOpenSerialDartu {929x}
\def \OpenroadTypicalSpeedupOpenParallelDartu {15x}
\def \OpenroadTypicalSpeedupSyntSerialDartu {112x}
\def \OpenroadTypicalSpeedupSyntParallelDartu {2x}
\def \OpenroadTypicalSpeedupOpenSerialNgspice {8,702x}
\def \OpenroadTypicalSpeedupOpenParallelNgspice {136x}
\def \OpenroadTypicalSpeedupSyntSerialNgspice {1,044x}
\def \OpenroadTypicalSpeedupSyntParallelNgspice {16x}

\def \OpenroadFastFailratio{0.07}                                
\def \OpenroadFastAllcasesAbserrorGnnceffMean {0.01}    
\def \OpenroadFastAllcasesAbserrorGnnceffMax {7.23}         
\def \OpenroadFastAllcasesAbserrorGnnceffStdev {0.07}       
\def \OpenroadFastAllcasesAbserrorDartuMean {0.02}          
\def \OpenroadFastAllcasesAbserrorDartuMax {21.87}      
\def \OpenroadFastAllcasesAbserrorDartuStdev {0.32}    
\def \OpenroadFastFailAbserrorGnnceffMean {0.68}        
\def \OpenroadFastFailAbserrorGnnceffMax {3.65}           
\def \OpenroadFastFailAbserrorGnnceffStdev {0.80}         
\def \OpenroadFastFailAbserrorDartuMean {9.92}            
\def \OpenroadFastFailAbserrorDartuMax {21.87}               
\def \OpenroadFastFailAbserrorDartuStdev {5.15}               
\def \OpenroadFastNofailAbserrorGnnceffMean {0.01}           
\def \OpenroadFastNofailAbserrorGnnceffMax {7.23}        
\def \OpenroadFastNofailAbserrorGnnceffStdev {0.07}      
\def \OpenroadFastNofailAbserrorDartuMean {0.01}          
\def \OpenroadFastNofailAbserrorDartuMax {11.19}           
\def \OpenroadFastNofailAbserrorDartuStdev {0.10}           
\def \OpenroadFastAllcasesAbserrorratioGnnceffMean {0.33}  
\def \OpenroadFastAllcasesAbserrorratioGnnceffMax {85.10}  
\def \OpenroadFastAllcasesAbserrorratioGnnceffStdev {1.48}  
\def \OpenroadFastAllcasesAbserrorratioDartuMean {0.60}     
\def \OpenroadFastAllcasesAbserrorratioDartuMax {609.76}
\def \OpenroadFastAllcasesAbserrorratioDartuStdev {7.40}  
\def \OpenroadFastFailAbserrorratioGnnceffMean {16.32}      
\def \OpenroadFastFailAbserrorratioGnnceffMax {85.10}      
\def \OpenroadFastFailAbserrorratioGnnceffStdev {19.45}   
\def \OpenroadFastFailAbserrorratioDartuMean {226.42}     
\def \OpenroadFastFailAbserrorratioDartuMax {609.76}         
\def \OpenroadFastFailAbserrorratioDartuStdev {130.55}       
\def \OpenroadFastNofailAbserrorratioGnnceffMean {0.32}       
\def \OpenroadFastNofailAbserrorratioGnnceffMax {83.86}  
\def \OpenroadFastNofailAbserrorratioGnnceffStdev {1.36}
\def \OpenroadFastNofailAbserrorratioDartuMean {0.45}     
\def \OpenroadFastNofailAbserrorratioDartuMax {609.76}     
\def \OpenroadFastNofailAbserrorratioDartuStdev {3.84}     
\def \OpenroadFastAllcasesDelayabserrorGnnceffMean {0.01}   
\def \OpenroadFastAllcasesDelayabserrorGnnceffMax {3.58}
\def \OpenroadFastAllcasesDelayabserrorGnnceffStdev {0.07}
\def \OpenroadFastAllcasesDelayabserrorDartuMean {0.02} 
\def \OpenroadFastAllcasesDelayabserrorDartuMax {30.41}
\def \OpenroadFastAllcasesDelayabserrorDartuStdev {0.31}
\def \OpenroadFastFailDelayabserrorGnnceffMean {0.61}      
\def \OpenroadFastFailDelayabserrorGnnceffMax {3.14}      
\def \OpenroadFastFailDelayabserrorGnnceffStdev {0.72}
\def \OpenroadFastFailDelayabserrorDartuMean {8.77}              
\def \OpenroadFastFailDelayabserrorDartuMax {30.41}
\def \OpenroadFastFailDelayabserrorDartuStdev {6.37}
\def \OpenroadFastNofailDelayabserrorGnnceffMean {0.01}
\def \OpenroadFastNofailDelayabserrorGnnceffMax {3.58}
\def \OpenroadFastNofailDelayabserrorGnnceffStdev {0.06}
\def \OpenroadFastNofailDelayabserrorDartuMean {0.02}
\def \OpenroadFastNofailDelayabserrorDartuMax {7.12}
\def \OpenroadFastNofailDelayabserrorDartuStdev {0.08}
\def \OpenroadFastAllcasesTimeGnnceffMean {49}
\def \OpenroadFastAllcasesTimeGnnceffMax {49}
\def \OpenroadFastAllcasesTimeGnnceffStdev {0}
\def \OpenroadFastFailTimeGnnceffMean {49}
\def \OpenroadFastFailTimeGnnceffMax {49}
\def \OpenroadFastFailTimeGnnceffStdev {0}
\def \OpenroadFastNofailTimeGnnceffMean {49}
\def \OpenroadFastNofailTimeGnnceffMax {49}
\def \OpenroadFastNofailTimeGnnceffStdev {0}
\def \OpenroadFastAllcasesTimeSyntSerialDartuMean {5,437}
\def \OpenroadFastAllcasesTimeSyntSerialDartuMax {121,189}
\def \OpenroadFastAllcasesTimeSyntSerialDartuStdev {1,822}
\def \OpenroadFastFailTimeSyntSerialDartuMean {6,653}
\def \OpenroadFastFailTimeSyntSerialDartuMax {61,363}
\def \OpenroadFastFailTimeSyntSerialDartuStdev {7,099}
\def \OpenroadFastNofailTimeSyntSerialDartuMean {5,436}
\def \OpenroadFastNofailTimeSyntSerialDartuMax {121,189}
\def \OpenroadFastNofailTimeSyntSerialDartuStdev {1,812}
\def \OpenroadFastAllcasesTimeOpenSerialDartuMean {45,309}
\def \OpenroadFastAllcasesTimeOpenSerialDartuMax {1,009,910}
\def \OpenroadFastAllcasesTimeOpenSerialDartuStdev {15,186}
\def \OpenroadFastFailTimeOpenSerialDartuMean {55,438}
\def \OpenroadFastFailTimeOpenSerialDartuMax {511,357}
\def \OpenroadFastFailTimeOpenSerialDartuStdev {59,156}
\def \OpenroadFastNofailTimeOpenSerialDartuMean {45,302}
\def \OpenroadFastNofailTimeOpenSerialDartuMax {1,009,910}
\def \OpenroadFastNofailTimeOpenSerialDartuStdev {15,099}
\def \OpenroadFastAllcasesTimeSyntParallelDartuMean {85}
\def \OpenroadFastAllcasesTimeSyntParallelDartuMax {1,894}
\def \OpenroadFastAllcasesTimeSyntParallelDartuStdev {28}
\def \OpenroadFastFailTimeSyntParallelDartuMean {104}
\def \OpenroadFastFailTimeSyntParallelDartuMax {959}
\def \OpenroadFastFailTimeSyntParallelDartuStdev {111}
\def \OpenroadFastNofailTimeSyntParallelDartuMean {85}
\def \OpenroadFastNofailTimeSyntParallelDartuMax {1,894}
\def \OpenroadFastNofailTimeSyntParallelDartuStdev {28}
\def \OpenroadFastAllcasesTimeOpenParallelDartuMean {708}
\def \OpenroadFastAllcasesTimeOpenParallelDartuMax {15,780}
\def \OpenroadFastAllcasesTimeOpenParallelDartuStdev {237}
\def \OpenroadFastFailTimeOpenParallelDartuMean {866}
\def \OpenroadFastFailTimeOpenParallelDartuMax {7,990}
\def \OpenroadFastFailTimeOpenParallelDartuStdev {924}
\def \OpenroadFastNofailTimeOpenParallelDartuMean {708}
\def \OpenroadFastNofailTimeOpenParallelDartuMax {15,780}
\def \OpenroadFastNofailTimeOpenParallelDartuStdev {236}
\def \OpenroadFastAllcasesTimeSyntSerialNgspiceMean {51,307}
\def \OpenroadFastAllcasesTimeSyntSerialNgspiceMax {136,996}
\def \OpenroadFastAllcasesTimeSyntSerialNgspiceStdev {7,709}
\def \OpenroadFastFailTimeSyntSerialNgspiceMean {51,678}
\def \OpenroadFastFailTimeSyntSerialNgspiceMax {63,874}
\def \OpenroadFastFailTimeSyntSerialNgspiceStdev {7,093}
\def \OpenroadFastNofailTimeSyntSerialNgspiceMean {51,307}
\def \OpenroadFastNofailTimeSyntSerialNgspiceMax {136,996}
\def \OpenroadFastNofailTimeSyntSerialNgspiceStdev {7,710}
\def \OpenroadFastAllcasesTimeOpenSerialNgspiceMean {427,560}
\def \OpenroadFastAllcasesTimeOpenSerialNgspiceMax {1,141,632}
\def \OpenroadFastAllcasesTimeOpenSerialNgspiceStdev {64,245}
\def \OpenroadFastFailTimeOpenSerialNgspiceMean {430,650}
\def \OpenroadFastFailTimeOpenSerialNgspiceMax {532,285}
\def \OpenroadFastFailTimeOpenSerialNgspiceStdev {59,110}
\def \OpenroadFastNofailTimeOpenSerialNgspiceMean {427,560}
\def \OpenroadFastNofailTimeOpenSerialNgspiceMax {1,141,632}
\def \OpenroadFastNofailTimeOpenSerialNgspiceStdev {64,248}
\def \OpenroadFastAllcasesTimeSyntParallelNgspiceMean {802}
\def \OpenroadFastAllcasesTimeSyntParallelNgspiceMax {2,141}
\def \OpenroadFastAllcasesTimeSyntParallelNgspiceStdev {120}
\def \OpenroadFastFailTimeSyntParallelNgspiceMean {807}
\def \OpenroadFastFailTimeSyntParallelNgspiceMax {998}
\def \OpenroadFastFailTimeSyntParallelNgspiceStdev {111}
\def \OpenroadFastNofailTimeSyntParallelNgspiceMean {802}
\def \OpenroadFastNofailTimeSyntParallelNgspiceMax {2,141}
\def \OpenroadFastNofailTimeSyntParallelNgspiceStdev {120}
\def \OpenroadFastAllcasesTimeOpenParallelNgspiceMean {6,681}
\def \OpenroadFastAllcasesTimeOpenParallelNgspiceMax {17,838}
\def \OpenroadFastAllcasesTimeOpenParallelNgspiceStdev {1,004}
\def \OpenroadFastFailTimeOpenParallelNgspiceMean {6,729}
\def \OpenroadFastFailTimeOpenParallelNgspiceMax {8,317}
\def \OpenroadFastFailTimeOpenParallelNgspiceStdev {924}
\def \OpenroadFastNofailTimeOpenParallelNgspiceMean {6,681}
\def \OpenroadFastNofailTimeOpenParallelNgspiceMax {17,838}
\def \OpenroadFastNofailTimeOpenParallelNgspiceStdev {1,004}
\def \OpenroadFastSpeedupOpenSerialDartu {916x}
\def \OpenroadFastSpeedupOpenParallelDartu {14x}
\def \OpenroadFastSpeedupSyntSerialDartu {110x}
\def \OpenroadFastSpeedupSyntParallelDartu {2x}
\def \OpenroadFastSpeedupOpenSerialNgspice {8,648x}
\def \OpenroadFastSpeedupOpenParallelNgspice {135x}
\def \OpenroadFastSpeedupSyntSerialNgspice {1,038x}
\def \OpenroadFastSpeedupSyntParallelNgspice {16x}

\def \OpenroadSlowFailratio{0.00}                                
\def \OpenroadSlowAllcasesAbserrorGnnceffMean {0.00}    
\def \OpenroadSlowAllcasesAbserrorGnnceffMax {7.31}         
\def \OpenroadSlowAllcasesAbserrorGnnceffStdev {0.06}       
\def \OpenroadSlowAllcasesAbserrorDartuMean {0.01}          
\def \OpenroadSlowAllcasesAbserrorDartuMax {14.37}      
\def \OpenroadSlowAllcasesAbserrorDartuStdev {0.10}    
\def \OpenroadSlowFailAbserrorGnnceffMean {0.97}        
\def \OpenroadSlowFailAbserrorGnnceffMax {2.24}           
\def \OpenroadSlowFailAbserrorGnnceffStdev {1.07}         
\def \OpenroadSlowFailAbserrorDartuMean {9.98}            
\def \OpenroadSlowFailAbserrorDartuMax {14.37}               
\def \OpenroadSlowFailAbserrorDartuStdev {5.86}               
\def \OpenroadSlowNofailAbserrorGnnceffMean {0.00}           
\def \OpenroadSlowNofailAbserrorGnnceffMax {7.31}        
\def \OpenroadSlowNofailAbserrorGnnceffStdev {0.06}      
\def \OpenroadSlowNofailAbserrorDartuMean {0.01}         
\def \OpenroadSlowNofailAbserrorDartuMax {9.60}            
\def \OpenroadSlowNofailAbserrorDartuStdev {0.07}           
\def \OpenroadSlowAllcasesAbserrorratioGnnceffMean {0.14}  
\def \OpenroadSlowAllcasesAbserrorratioGnnceffMax {87.35}  
\def \OpenroadSlowAllcasesAbserrorratioGnnceffStdev {0.69}  
\def \OpenroadSlowAllcasesAbserrorratioDartuMean {0.21}     
\def \OpenroadSlowAllcasesAbserrorratioDartuMax {220.81}
\def \OpenroadSlowAllcasesAbserrorratioDartuStdev {1.44}  
\def \OpenroadSlowFailAbserrorratioGnnceffMean {14.49}    
\def \OpenroadSlowFailAbserrorratioGnnceffMax {32.73}      
\def \OpenroadSlowFailAbserrorratioGnnceffStdev {15.07}   
\def \OpenroadSlowFailAbserrorratioDartuMean {155.48}     
\def \OpenroadSlowFailAbserrorratioDartuMax {220.81}         
\def \OpenroadSlowFailAbserrorratioDartuStdev {88.19}        
\def \OpenroadSlowNofailAbserrorratioGnnceffMean {0.14}       
\def \OpenroadSlowNofailAbserrorratioGnnceffMax {87.35}  
\def \OpenroadSlowNofailAbserrorratioGnnceffStdev {0.67}
\def \OpenroadSlowNofailAbserrorratioDartuMean {0.21}   
\def \OpenroadSlowNofailAbserrorratioDartuMax {192.25}     
\def \OpenroadSlowNofailAbserrorratioDartuStdev {0.96}     
\def \OpenroadSlowAllcasesDelayabserrorGnnceffMean {0.01}   
\def \OpenroadSlowAllcasesDelayabserrorGnnceffMax {21.58}
\def \OpenroadSlowAllcasesDelayabserrorGnnceffStdev {0.13}
\def \OpenroadSlowAllcasesDelayabserrorDartuMean {0.02}
\def \OpenroadSlowAllcasesDelayabserrorDartuMax {20.52}
\def \OpenroadSlowAllcasesDelayabserrorDartuStdev {0.15} 
\def \OpenroadSlowFailDelayabserrorGnnceffMean {1.41}     
\def \OpenroadSlowFailDelayabserrorGnnceffMax {3.36}      
\def \OpenroadSlowFailDelayabserrorGnnceffStdev {1.54}
\def \OpenroadSlowFailDelayabserrorDartuMean {14.62}             
\def \OpenroadSlowFailDelayabserrorDartuMax {20.52}
\def \OpenroadSlowFailDelayabserrorDartuStdev {8.45}
\def \OpenroadSlowNofailDelayabserrorGnnceffMean {0.01}
\def \OpenroadSlowNofailDelayabserrorGnnceffMax {21.58}
\def \OpenroadSlowNofailDelayabserrorGnnceffStdev {0.13}
\def \OpenroadSlowNofailDelayabserrorDartuMean {0.02}
\def \OpenroadSlowNofailDelayabserrorDartuMax {14.83}
\def \OpenroadSlowNofailDelayabserrorDartuStdev {0.11}
\def \OpenroadSlowAllcasesTimeGnnceffMean {49}
\def \OpenroadSlowAllcasesTimeGnnceffMax {49}
\def \OpenroadSlowAllcasesTimeGnnceffStdev {0}
\def \OpenroadSlowFailTimeGnnceffMean {49}
\def \OpenroadSlowFailTimeGnnceffMax {49}
\def \OpenroadSlowFailTimeGnnceffStdev {0}
\def \OpenroadSlowNofailTimeGnnceffMean {49}
\def \OpenroadSlowNofailTimeGnnceffMax {49}
\def \OpenroadSlowNofailTimeGnnceffStdev {0}
\def \OpenroadSlowAllcasesTimeSyntSerialDartuMean {5,522}
\def \OpenroadSlowAllcasesTimeSyntSerialDartuMax {32,947}
\def \OpenroadSlowAllcasesTimeSyntSerialDartuStdev {1,601}
\def \OpenroadSlowFailTimeSyntSerialDartuMean {5,441}
\def \OpenroadSlowFailTimeSyntSerialDartuMax {6,949}
\def \OpenroadSlowFailTimeSyntSerialDartuStdev {945}
\def \OpenroadSlowNofailTimeSyntSerialDartuMean {5,522}
\def \OpenroadSlowNofailTimeSyntSerialDartuMax {32,947}
\def \OpenroadSlowNofailTimeSyntSerialDartuStdev {1,601}
\def \OpenroadSlowAllcasesTimeOpenSerialDartuMean {46,015}
\def \OpenroadSlowAllcasesTimeOpenSerialDartuMax {274,560}
\def \OpenroadSlowAllcasesTimeOpenSerialDartuStdev {13,345}
\def \OpenroadSlowFailTimeOpenSerialDartuMean {45,339}
\def \OpenroadSlowFailTimeOpenSerialDartuMax {57,911}
\def \OpenroadSlowFailTimeOpenSerialDartuStdev {7,874}
\def \OpenroadSlowNofailTimeOpenSerialDartuMean {46,015}
\def \OpenroadSlowNofailTimeOpenSerialDartuMax {274,560}
\def \OpenroadSlowNofailTimeOpenSerialDartuStdev {13,345}
\def \OpenroadSlowAllcasesTimeSyntParallelDartuMean {86}
\def \OpenroadSlowAllcasesTimeSyntParallelDartuMax {515}
\def \OpenroadSlowAllcasesTimeSyntParallelDartuStdev {25}
\def \OpenroadSlowFailTimeSyntParallelDartuMean {85}
\def \OpenroadSlowFailTimeSyntParallelDartuMax {109}
\def \OpenroadSlowFailTimeSyntParallelDartuStdev {15}
\def \OpenroadSlowNofailTimeSyntParallelDartuMean {86}
\def \OpenroadSlowNofailTimeSyntParallelDartuMax {515}
\def \OpenroadSlowNofailTimeSyntParallelDartuStdev {25}
\def \OpenroadSlowAllcasesTimeOpenParallelDartuMean {719}
\def \OpenroadSlowAllcasesTimeOpenParallelDartuMax {4,290}
\def \OpenroadSlowAllcasesTimeOpenParallelDartuStdev {209}
\def \OpenroadSlowFailTimeOpenParallelDartuMean {708}
\def \OpenroadSlowFailTimeOpenParallelDartuMax {905}
\def \OpenroadSlowFailTimeOpenParallelDartuStdev {123}
\def \OpenroadSlowNofailTimeOpenParallelDartuMean {719}
\def \OpenroadSlowNofailTimeOpenParallelDartuMax {4,290}
\def \OpenroadSlowNofailTimeOpenParallelDartuStdev {209}
\def \OpenroadSlowAllcasesTimeSyntSerialNgspiceMean {51,066}
\def \OpenroadSlowAllcasesTimeSyntSerialNgspiceMax {120,727}
\def \OpenroadSlowAllcasesTimeSyntSerialNgspiceStdev {7,692}
\def \OpenroadSlowFailTimeSyntSerialNgspiceMean {47,759}
\def \OpenroadSlowFailTimeSyntSerialNgspiceMax {49,438}
\def \OpenroadSlowFailTimeSyntSerialNgspiceStdev {1,988}
\def \OpenroadSlowNofailTimeSyntSerialNgspiceMean {51,067}
\def \OpenroadSlowNofailTimeSyntSerialNgspiceMax {120,727}
\def \OpenroadSlowNofailTimeSyntSerialNgspiceStdev {7,692}
\def \OpenroadSlowAllcasesTimeOpenSerialNgspiceMean {425,554}
\def \OpenroadSlowAllcasesTimeOpenSerialNgspiceMax {1,006,056}
\def \OpenroadSlowAllcasesTimeOpenSerialNgspiceStdev {64,096}
\def \OpenroadSlowFailTimeOpenSerialNgspiceMean {397,995}
\def \OpenroadSlowFailTimeOpenSerialNgspiceMax {411,982}
\def \OpenroadSlowFailTimeOpenSerialNgspiceStdev {16,566}
\def \OpenroadSlowNofailTimeOpenSerialNgspiceMean {425,555}
\def \OpenroadSlowNofailTimeOpenSerialNgspiceMax {1,006,056}
\def \OpenroadSlowNofailTimeOpenSerialNgspiceStdev {64,097}
\def \OpenroadSlowAllcasesTimeSyntParallelNgspiceMean {798}
\def \OpenroadSlowAllcasesTimeSyntParallelNgspiceMax {1,886}
\def \OpenroadSlowAllcasesTimeSyntParallelNgspiceStdev {120}
\def \OpenroadSlowFailTimeSyntParallelNgspiceMean {746}
\def \OpenroadSlowFailTimeSyntParallelNgspiceMax {772}
\def \OpenroadSlowFailTimeSyntParallelNgspiceStdev {31}
\def \OpenroadSlowNofailTimeSyntParallelNgspiceMean {798}
\def \OpenroadSlowNofailTimeSyntParallelNgspiceMax {1,886}
\def \OpenroadSlowNofailTimeSyntParallelNgspiceStdev {120}
\def \OpenroadSlowAllcasesTimeOpenParallelNgspiceMean {6,649}
\def \OpenroadSlowAllcasesTimeOpenParallelNgspiceMax {15,720}
\def \OpenroadSlowAllcasesTimeOpenParallelNgspiceStdev {1,002}
\def \OpenroadSlowFailTimeOpenParallelNgspiceMean {6,219}
\def \OpenroadSlowFailTimeOpenParallelNgspiceMax {6,437}
\def \OpenroadSlowFailTimeOpenParallelNgspiceStdev {259}
\def \OpenroadSlowNofailTimeOpenParallelNgspiceMean {6,649}
\def \OpenroadSlowNofailTimeOpenParallelNgspiceMax {15,720}
\def \OpenroadSlowNofailTimeOpenParallelNgspiceStdev {1,002}
\def \OpenroadSlowSpeedupOpenSerialDartu {930x}
\def \OpenroadSlowSpeedupOpenParallelDartu {15x}
\def \OpenroadSlowSpeedupSyntSerialDartu {112x}
\def \OpenroadSlowSpeedupSyntParallelDartu {2x}
\def \OpenroadSlowSpeedupOpenSerialNgspice {8,599x}
\def \OpenroadSlowSpeedupOpenParallelNgspice {134x}
\def \OpenroadSlowSpeedupSyntSerialNgspice {1,032x}
\def \OpenroadSlowSpeedupSyntParallelNgspice {16x}

%% file: intro.tex
\section{Introduction}
\label{sec:intro}

Very-Large-Scale Integration (VLSI) ``physical synthesis'' design flows have interdependent stages of placement, routing, and static timing analysis (STA). The overall power, performance, area (PPA) metrics of a design depend on the routing, and yet routing performance depends on placement. STA has the crucial role of evaluating these results both during design optimization and sign-off.

Gate delay and effective capacitance (\ceff) computation are critical in STA because they directly determine the accuracy and reliability of delay and slew estimates. Accurate gate delay modeling, which depends on \ceff{}, is essential for sign-off timing, where small inaccuracies can lead to timing violations in silicon. Incremental STA further relies on quickly updating delays after localized design changes, which demands fast yet accurate \ceff{} and delay recalculation. Moreover, with modern processes exhibiting significant variation, STA must be performed across multiple process-voltage-temperature (PVT) corners and statistical scenarios to ensure robustness. In such cases, precise \ceff{} and delay modeling across all corners is crucial to verify timing correctness under all manufacturing and environmental conditions.

\ceff{} computation is also challenging because it depends on the design, library, and technology parameters. For example, the routing topology, the layers used, the fanout gates, the cross-coupling capacitance as well as the driving characteristics of the source gate itself all have an impact on \ceff.

The state-of-the-art solutions tend to simplify \ceff{} modeling into several discrete steps of interconnect Model Order Reduction (MOR), \ceff{} computation, and gate delay computation. MOR is necessary for computing interconnect delay by capturing distributed parasitic effects with fast, iterative solutions and scaling to large designs, but is also used to simplify interconnect for gate delay models. Gate delay models are based on either Non-Linear Delay Models (NLDM) or Composite Current Source (CCS). For NLDM models, \ceff{} is critical for utilizing the 2-dimensional NLDM tables, indexed by capacitive load and input slew, to find accurate delay and slew of gates. For CCS models, \ceff{} plays a different, but still important role of a first-order approximation to avoid full CCS evaluation, to help simplify RC networks, and to analyze sensitivity across multiple corners~\cite{tempus,primetime}. Together, MOR and gate-delay models have been combined into iterative approaches~\cite{dmp} that compute gate delays considering gates that drive distributed interconnects. 

Our approach combines MOR and \ceff{} computation into a single Graph Neural Network (GNN) model that provides both more accurate computation and parallelization compared to current algorithms. In particular, we propose
\begin{itemize}
    \item The first GNN model for post-layout \ceff{} computation.
    \item Implicit learnable MOR with graph pooling. 
    \item A training methodology using synthetic data.
    \item Hyperparameter optimization of the model.
    \item Evaluation on real-world multi-corner benchmarks using a complete open-source RTL2GDS tool flow.
\end{itemize}

%% file: background.tex
\section{Background}
\label{sec:background}

\subsection{Model Order Reduction}

\begin{figure}[htb]
    \centering
    \includegraphics[width=0.8\linewidth]{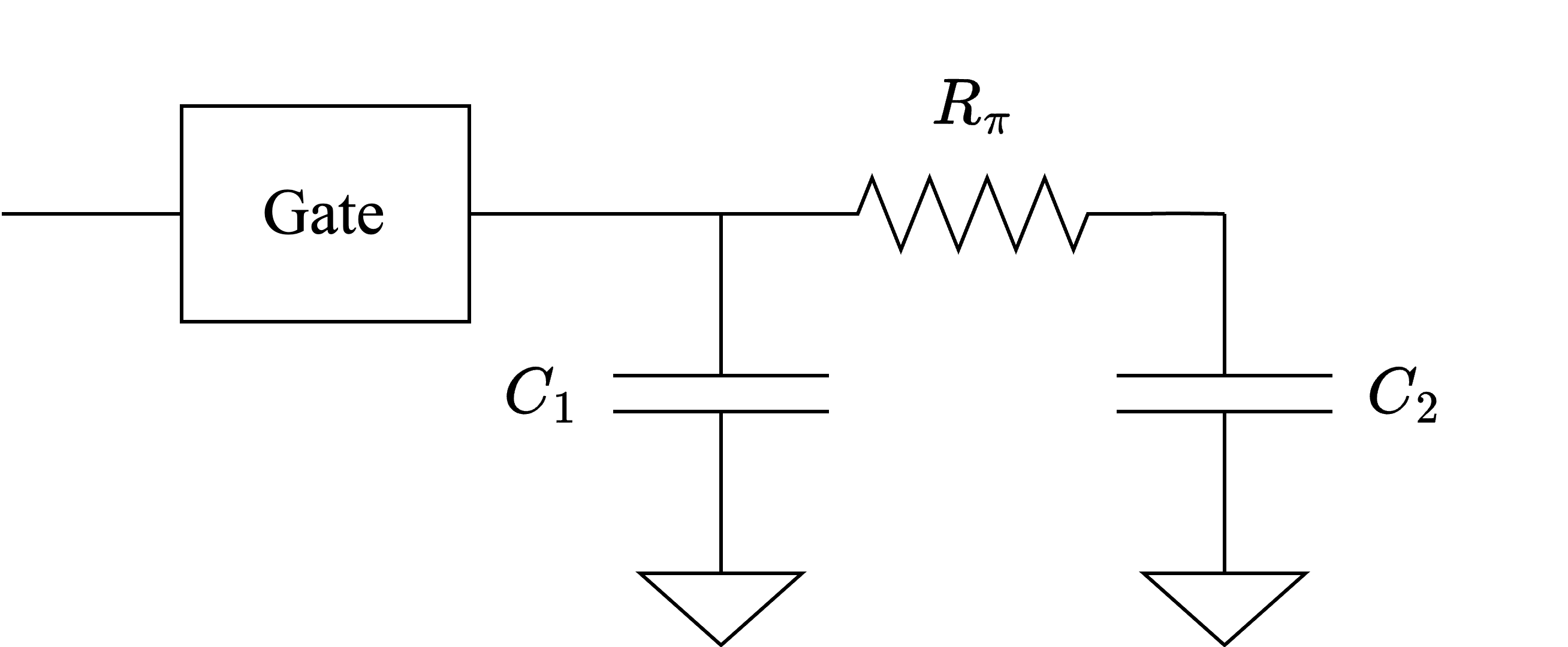}
    \caption{The \pimodel{} is used for modeling driving point admittance in gate delay computation.}
    \label{fig:pi_model}
    \vspace{-0.5cm}
\end{figure}

In most Static Timing Analysis (STA) approaches, delay estimation models are utilized due to the computational complexity involved in simulating the delay of gates and wires. However, creating accurate estimation models for arbitrary interconnect structures is a challenging task. To address this issue, a simplified model known as ``\pimodel{}'' is proposed by O'Brien et al.~\cite{pi_model} as illustrated in Fig.~\ref{fig:pi_model}. This model reduction technique leverages an RC network, which captures the parasitic resistance and capacitance of the interconnect, to approximate the driving-point admittance characteristics. Subsequently, the interconnect RC network is condensed into three key parameters: $C_1$, $C_2$, and $R_\pi$. $C_1$ represents the source capacitance. The downstream capacitance is clustered into a single value ($C_2$), with resistance ($R_\pi$) in between $C_1$ and $C_2$ shielding the downstream capacitance.

\subsection{Delay Models}

Non-Linear Delay Models (NLDM) use 2-dimensional tables indexed by the capacitive load and input slew of gates. These tables are constructed for each gate in the technology by prerunning a number of simulations sampled from the 2-dimensional plane of capacitive load and input slew. For values not perfectly overlapping the table indices, NLDM-based methods interpolate these values to statistically model the output delay and slew of gates. The problem with NLDM is the trade-off between the size of tables and the statistical interpolation errors. 

Composite Current Source (CCS) models, on the other hand, characterize the RC network as a function of time and return its current waveform. Therefore, CCS-based methods provide more accurate modeling of gate delay and slew, though the downside is the slowness of current waveform calculation. CCS still performs MOR and the delay modeling with the CCS tables over time by using the input and output voltage to predict the current at given times.

\subsection{Effective Capacitance Methodology}
The delay of a logic gate depends on both the capacitance of the interconnect and the fanout gate pins. The lumped total capacitance (\ctotal) of these, however, becomes pessimistic as the size of transistors becomes smaller and processes become faster. In modern technologies, gates do not see the \ctotal{} of the nets due to the resistive shielding effect of the large RC interconnects, which means that STA using \ctotal{} is too pessimistic. Instead, effective capacitance (\ceff) modeling accounts for the resistive shielding effect in STA~\cite{ceff} while also keeping delay models, such as NLDM, to small numbers of dimensions.

Modern STA tools use the heuristic of Dartu et al.~\cite{dmp}, or variations of it, combined with the O'Brien et al. method to compute the \ceff{} of a net. This heuristic uses the \pimodel{} of the net obtained from the O'Brien algorithm and starts with an initial \ceff{} guess equal to the \ctotal{}. The Dartu algorithm computes the average currents drawn from the source by the \pimodel{} and the \ceff{} model and iteratively updates \ceff{} to satisfy $I_{\pi}=I_{C\text{eff}}$. The converged value of \ceff{} indexes the gate delay model for gate delay estimation. This heuristic makes several assumptions including that: the driving gate has a fixed resistance, the reduced \pimodel{} is available, pre-characterized timing tables are available, and that slews are simple ramps.

\subsection{Graph Neural Networks}
Graph Neural Networks (GNNs) are a machine learning technique similar to the well-known Convolutional Neural Networks (CNNs); however, they take general graphs with varying node and edge degrees as input~\cite{gcn,gat,gatv2,graphsage}. Technically, GNNs are generalized CNNs with CNNs being a special case of GNNs with fixed graph structures. GNN models interpret relations between nodes of a given graph and can be trained for classification and regression tasks on nodes, edges, or the graph.

The core principle behind GNNs is called ``message passing'' where the features of nodes are updated using learnable weights and the features. That is, each GNN layer can use the information from neighbors that are one step away to update the feature embeddings of a node. Multiple message-passing layers enable nodes to eventually learn about further away nodes and features.

Graph Attention Networks (GATs)~\cite{gat,gatv2} are specialized variants of GNNs, where an additional attention mechanism can learn the relations of neighbors~\cite{attention}. For each neighbor of a node, GATs add an attention coefficient based on weighted feature similarity. All neighbor coefficients of a node are then compared using the softmax function at each layer. Multiple attention heads can also be incorporated so that GATs can understand how different weightings of neighbor features may affect the output of that node differently~\cite{gat}.

Pooling techniques have been used to reduce the size of an input matrix by combining features of pooled pixels and GNNs, just like CNNs, have their own pooling techniques. Graph pooling (GP) layers reduce the size of an input graph, while also providing a broader perspective for graph-level inference. Global GP takes the sum, maximum, or average of each feature in all nodes so that we end up with a single node, which is just a single tensor of features. This tensor can then be fed to a multi-layer perceptron (MLP) to perform graph-level classification or regression. Another promising technique of GP is attentional aggregation~\cite{attentional_aggregation}, which pools each feature in all nodes, similar to global GP. However, it passes the feature channels through a weight vector so that the relative importance of each node feature also becomes learnable. Therefore, an attentional aggregation layer can learn a more sophisticated function of global pooling.



\subsection{Related Work}
In recent years, there have been many works in both academia and the industry towards machine learning-based timing analysis methods. For example, He et al.~\cite{rw_2} proposed the HGATTrans model that predicts the timing of a net based on pre-routing information. Although the model does not produce highly accurate predictions, it gives insights about a net's performance post-routing, also showing the potential speedup gain of graphics processing unit (GPU) parallelization. Chhabria et al.~\cite{rw_4} proposed three models to predict wire delay, wire slew, and the \pimodel{} based on post-global routing results. Some works in the literature focused on RC interconnect delay, such as the ResNet-based neural network model of Liu et al.~\cite{rw_1}. Their model predicts the path delay of a given RC interconnect using features such as cell delay, Elmore delay, and \ceff. Zhu et al.~\cite{rw_3} and Hu et al.~\cite{rw_5} proposed GNN-based models for predicting the delay from the driver node to each fanout node. While these previous works focused on interconnect delay, there are also papers that examine parasitic interconnect capacitance. For instance, Ren et al.~\cite{rw_6} and Shen et al.~\cite{rw_7} proposed GNN-based methods for pre-layout parasitic feature prediction, including \ceff{}. Although these models can provide useful information about the performance of nets to be placed and routed in advance, their accuracies are not good enough for critical post-layout and sign-off timing. Our proposed method is the first in critical post-layout \ceff{} prediction using a GNN.

%% file: proposed.tex
\section{Proposed \ceff{} Modeling}
\label{sec:model}

\subsection{Graph Representation}

Each unreduced, extracted RC network is a graph, \rcnetwork{}, where the nodes are pins and junctions, and the edges are wires connecting the pins. Both the driver pin and fanout gate pins are modeled as nodes. Hence, using \rcnetwork{} in GNNs directly is possible. However, our experiments have shown that edge features are not as effective as node features since most GNN convolution implementations are ``node centric'' and often do not support edge features directly. We observed that trained models directly using this graph had difficulties learning the task of modeling \ceff{}.

Consequently, our input GNN graphs, $G_\text{GNN}=(\mathcal{V}, \mathcal{E})$, represent \rcnetwork{} with two types of nodes: node-node (\nodenode{}) and edge-node (\edgenode{}). This means that not only are the circuit nodes represented by graph nodes, but also the wires (i.e., the edges) are graph nodes. This representation of a \rcnetwork{} increases the number of nodes from $|\mathcal{V}_\text{RC}|=n$ to $|\mathcal{V}_\text{GNN}|=2n-1$, therefore increasing the depth of the graph. However, most of the \nodenode{} can be trimmed, as explained later in this section. For this purpose, we make a distinction between \nodenode{} that are pins (\nodenodepin{}), and \nodenode{} that are junction nodes (\nodenodejunc{}).

The features needed to compute the \ceff{}, according to the prior O'Brien/Dartu heuristic, are the low voltage threshold (\vl{}) to high voltage threshold (\vh{}) input slew of the driver gate (\slew), resistance of the driver gate (\rd), resistance of wire segments (\wireres), capacitance of wire segments (\wirecap), and capacitance of fanout gate pins (\pincap). Although the Dartu algorithm uses the reduced-order \pimodel{} to represent the RC interconnect graph of a net, we use the RC interconnect itself without any reduction since our method has implicit order reduction using attentional aggregation. If a net has cross-coupling capacitance, we easily include it by connecting a virtual \nodenodepin{} node to the \edgenode{} of the wire segment that the coupling capacitance belongs to. Since a \rcnetwork{} has all of these necessary features, we can construct its \gnngraph{}.

As shown in Table~\ref{table:features}, the \nodenodepin{} have \rd{}, \slew{}, or \pincap{} features, depending on their direction type. The \edgenode{} have \wireres{} and \wirecap{} features. We also added a few new features to help the model's receptive field during message passing. The first such feature is a one-hot encoded \nodenodepin{} direction type flag to make the model better understand which nodes are driver (\driverflag{}) and fanout (\fanoutflag{}) pins. The second and third such features are the cumulative upstream resistance (\upres) and downstream capacitance (\downcap) of all nodes which correspond to well-known models such as Elmore delay. With these additional features, a node has the information of the total resistance down from the driver node and the total capacitance up from the fanout nodes. The fourth such feature is the number of hops from the driver node (\numhops) which helps to model the direction toward or away from the driver or fanout nodes.

\begin{figure}[htb]
    \centering
    \includegraphics[width=0.95\linewidth]{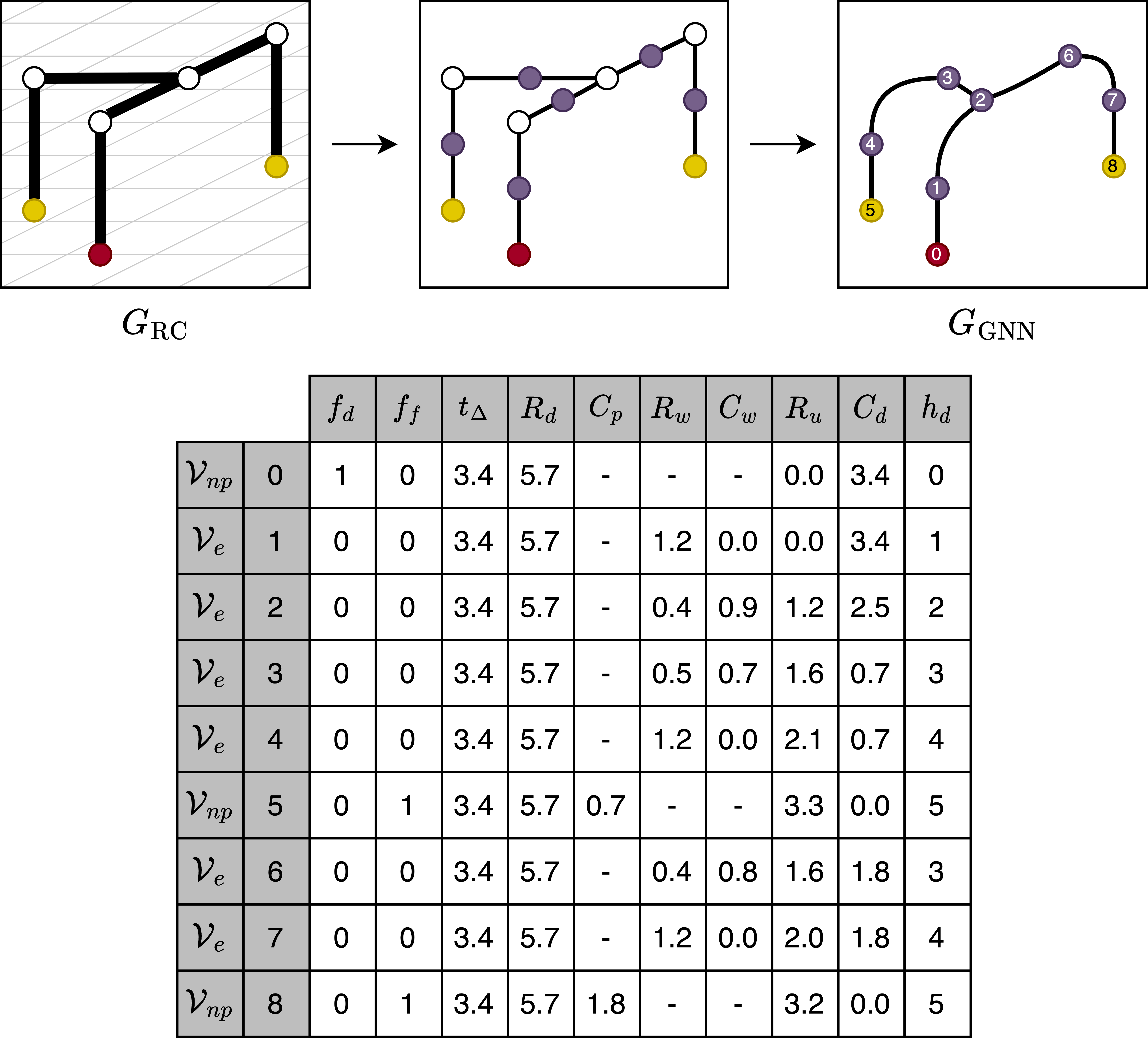}
    \caption{Simple example of a \gnngraph{} shows the one-to-one mapping of our graphs to physical wires.}
    \label{fig:toy_example}
\end{figure}

An astute observer might notice from these feature descriptions that the set of \nodenodejunc{} does not actually have any meaningful features. Thus, we can process these graphs to remove all \nodenodejunc{} to shrink the size of the graph, which helps the training performance and inference speed. All \nodenodejunc{} are, therefore, removed, and their neighbors are connected in the direction current flows from the driver to fanout nodes to keep the graph intact and still have the representation of physical connectivity. As an example in Fig.~\ref{fig:toy_example}, node 2 is directly connected to both nodes 3 and 6 when the intermediate junction node is removed.

We assign \rd{} and \slew{} to all nodes as global features since these features affect the whole graph.  The one-to-one mapping of a \rcnetwork{} to its \gnngraph{} is illustrated in Fig.~\ref{fig:toy_example}, and a full listing of example features is in Table~\ref{table:features}.

\begin{table}[htb]
\centering
\def\arraystretch{1.5}
\caption{Node-Level Features}
\label{table:features}
\centering
\setlength\tabcolsep{5pt}
\begin{tabular}{|c|c|c|}
\hline
Feature Name & Notation & Description \\
\hline
\multicolumn{1}{|l|}{Driver flag} & \driverflag{} & \multicolumn{1}{l|}{If the node is the driver pin} \\
\hline
\multicolumn{1}{|l|}{Fanout flag} & \fanoutflag{} & \multicolumn{1}{l|}{If the node is a fanout pin} \\
\hline
\multicolumn{1}{|l|}{Input slew} & \slew{} & \multicolumn{1}{l|}{\vl{}-\vh{} input slew of the driver} \\
\hline
\multicolumn{1}{|l|}{Driver resistance} & \rd{} & \multicolumn{1}{l|}{Resistance of the driver gate} \\
\hline
\multicolumn{1}{|l|}{Pin capacitance} & \pincap{} & \multicolumn{1}{l|}{Capacitance of a fanout pin} \\
\hline
\multicolumn{1}{|l|}{Wire resistance} & \wireres{} & \multicolumn{1}{l|}{Resistance of a wire segment} \\
\hline
\multicolumn{1}{|l|}{Wire capacitance} & \wirecap{} & \multicolumn{1}{l|}{Capacitance of a wire segment} \\
\hline
\multicolumn{1}{|l|}{Upstream resistance} & \upres{} & \multicolumn{1}{l|}{Total resistance down to a node} \\
\hline
\multicolumn{1}{|l|}{Downstream capacitance} & \downcap{} & \multicolumn{1}{l|}{Total capacitance up to a node} \\
\hline
\multicolumn{1}{|l|}{Number of hops} & \numhops{} & \multicolumn{1}{l|}{Number of hops from the driver} \\
\hline
\end{tabular}
\end{table}

The label of each \gnngraph{} is the ratio of $C_\text{eff}/C_\text{total}$ since our experiments showed that the models can model the \ceff{} more accurately this way. We believe this is due to the ratio being normalized and better learned than an absolute value. As the \ctotal{} of a net is known and easily calculated (and present on the driver node as an added feature), this normalization does not have any negative impact on performance. It does add the benefit of preventing the model from computing $C_\text{eff}<0$ or $C_\text{eff}>C_\text{total}$, which are not physically possible.

\begin{figure*}[tb] 
\vspace{-0.6cm}
\begin{center}
\includegraphics[width=\textwidth]{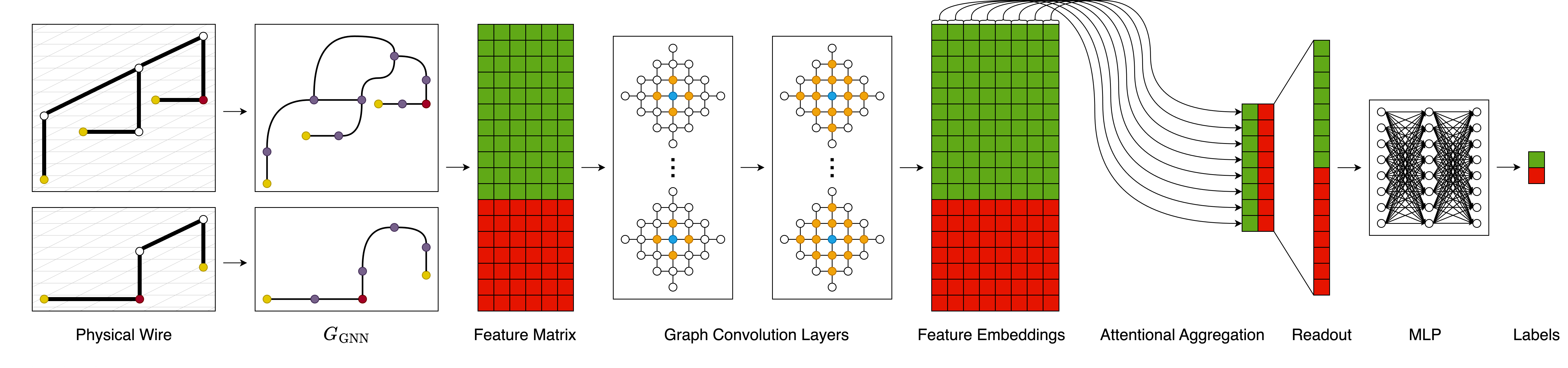}
 \caption {The model flow starts by converting the physical wires of nets into graphs. Feature vectors and edge lists are constructed. The model processes the inputs into feature embeddings for a number of GAT layers using multiple attention heads and the ELU activation function. The outputs of convolution layers are flattened with the attentional aggregation layer and given to the MLP, which computes the final labels.
 \label{fig:model_structure}}
 \end{center}
\vspace{-0.6cm}
\end{figure*}

\subsection{Testing Data Retrieval}

We use OpenROAD~\cite{openroad} for the designs provided in OpenROAD-flow-scripts~\cite{openroadflowscripts} with the ASAP7 technology~\cite{asap7} to produce testing data. OpenROAD's STA tool, OpenSTA~\cite{opensta}, uses the OpenRCX~\cite{openrcx} parasitic extractor to represent physical wires as \rcnetwork{}. We implemented additional helper functions in OpenROAD and OpenSTA to be able to extract the aforementioned necessary features in OpenROAD's Python API. 

After all features are received from the signed-off designs, each \rcnetwork{} is dumped as a SPICE file. We run timing simulations on the SPICE files using the Ngspice circuit simulator~\cite{ngspice}. The graph-level labels are obtained using a binary search of the simulated driver gate output delay to match the \ceff~delay to the RC network delay, similar to how the O'Brien/Dartu heuristic computes \ceff{} as described in Section~\ref{sec:background}.

In order to evaluate our method on different corners, we performed this process for the typical, fast, and slow corners of ASAP7. We call these datasets \openroadtypical{}, \openroadfast{}, and \openroadslow{} respectively. All these datasets have the same nets, but each corner has different \rd{}, \slew{}, and \pincap{} values depending on the PVT.

\subsection{Training Data Generation}

\begin{figure}[htb]
    \centering
    \includegraphics[width=\linewidth]{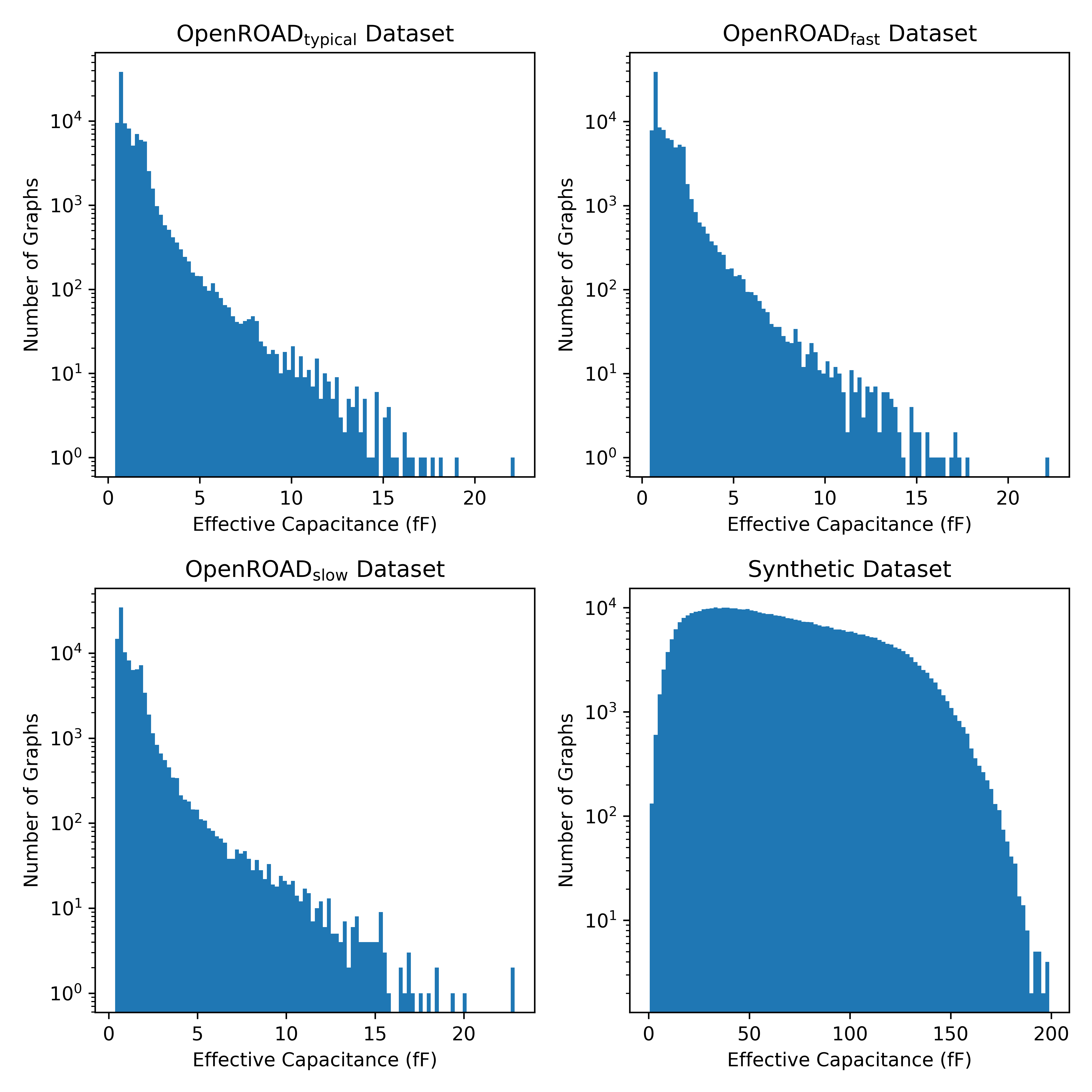}
    \caption{\ceff~distribution of all datasets show that the OpenROAD datasets' \ceff{} values for all corners are not uniform across the range, but the synthetic dataset's labels are more uniform and cover more \ceff{} cases.}
    \label{fig:openroad_ceff_distribution}
    \vspace{-0.3cm}
\end{figure}

After experimenting with all the retrieved data and based on our past experiences in machine learning, we decided to create a randomly generated, uniformly distributed training dataset. This is due to the majority of physical nets in the actual designs being small. Hence, the OpenROAD datasets have very skewed distributions of \ceff, as seen in Fig.~\ref{fig:openroad_ceff_distribution}, which biases learning of small nets more than larger nets which may, in fact, be more challenging to model. To make our model generalize better in all cases, we used more uniformly distributed training and testing data.

We used the GeoSteiner tool~\cite{geosteiner} to generate synthetic rectilinear Steiner minimum trees (RSMTs) with coordinates chosen randomly between 0 and 1,000,000 as integers which are then scaled to bounding boxes between 30nm and 100,000nm on the longer side. We generated 10,000 such nets for each degree between 3 and 50. After creating the synthetic RSMTs, we extracted the technology layers of the ASAP7 technology and assigned layers to the nodes and edges of RSMTs. For \slew{}, \rd{}, and \pincap{}, we selected the range of these features from the \openroadtypical{} dataset to be randomly sampled.  This synthetic dataset makes the learning process more uniform across both net degrees as shown in Table~\ref{table:training_dataset_stats} and \ceff{} as shown in Fig.~\ref{fig:openroad_ceff_distribution}. 

The synthetic dataset is converted to \rcnetwork{} and dumped as SPICE files. The labels are obtained using the same method as the OpenROAD datasets. A random 10\% of \gnngraph{} from each net degree of the synthetic dataset are used for training, while the rest are used for testing. Real-life testing is done on the real net distributions extracted from the OpenROAD designs which were not used for training.

\begin{table}[htb]
\centering
\def\arraystretch{1.5}
\caption{Synthetic Training Dataset Statistics\\(Note, \# of nets are in thousands.)}
\label{table:training_dataset_stats}
\centering
\setlength\tabcolsep{5.5pt}
\begin{tabular}{|c|c|c|c|c|c|c|c|}
\hline
\multirow{2}{*}{\shortstack{\# of nets\\($\times10^3$)}}& \multicolumn{6}{c|}{\# of nets by degree ($\times10^3$)} & \multirow{2}{*}{\shortstack{Degree of\\largest net}} \\
\cline{2-7}
     & 3-9 & 10-19 & 20-29 & 30-39 & 40-49 & 50 & \\
\hline
48  & 7   & 10    & 10    & 10    & 10    & 1         & 50 \\
\hline
\end{tabular}
\vspace{-0.3cm}
\end{table}

\subsection{Proposed Model and Hyperparameters}
We propose the GNN-\ceff{} model that is illustrated in Fig.~\ref{fig:model_structure}. 
Our GNN model comprises a number of \modelConvLayer{} convolutional layers~\modelConvLayerRef{}, followed by an attentional aggregation layer~\cite{attentional_aggregation} that is then fed to a multi-layer perceptron (MLP). We adopt the supervised training methodology using the labels obtained from SPICE simulations. We used the Ray Tune framework~\cite{raytune} to explore the hyperparameter space shown in Table~\ref{table:hyperparameters}.

We utilized layer and attention dropout methods for the training of our model. The optimal dropout rate is small because a high dropout rate leaves many \gnngraph{} disconnected, thus harming the learning process. However, a small enough dropout rate makes the training less prone to over-fitting.


\begin{table}[ht]
\vspace{-0.1cm}
\centering
\def\arraystretch{1.5}
\caption{Model Hyperparameters}
\label{table:hyperparameters}
\setlength\tabcolsep{3pt}
\begin{tabular}{|l|c|c|}
\hline
\multicolumn{1}{|c|}{Parameter} & Range Explored & Best Value \\
\hline
\multirow{4}{*}{Convolution layer} & GCN \cite{gcn} & \multirow{4}{*}{\modelConvLayer{} \modelConvLayerRef{}} \\
& GAT \cite{gat} &  \\
& GATv2 \cite{gatv2} &  \\
& GraphSage \cite{graphsage} &  \\
\hline
Number of convolution layers     & [2, 6]      & \modelNumConvLayers{} \\
\hline
Number of convolution channels   & [16, 256]   & \modelNumConvChannels{} \\
\hline
Number of attention heads        & [2, 16]     & \modelNumAttentionHeads{} \\
\hline
Number of linear layers          & [2, 6]      & \modelNumLinearLayers{} \\
\hline
Number of linear channels        & [16, 256]   & \modelNumLinearChannels{} \\
\hline
Layer dropout                    & [0.0, 0.25] & \modelLayerDropout{} \\
\hline
Attention dropout                & [0.0, 0.25] & \modelAttentionDropout{} \\
\hline
\end{tabular}
\vspace{-0.4cm}
\end{table}

%% file: results.tex
\section{Results and Discussion}
\label{sec:results}

\begin{table}[tb]
\centering
\def\arraystretch{1.5}
\caption{Test Datasets' Statistics\\(Note, \# of nets are in thousands.)}
\label{table:evaluation_dataset_stats}
\centering
\setlength\tabcolsep{2pt}
\begin{tabular}{|c|c|c|c|c|c|c|c|c|}
\hline
\multirow{2}{*}{Dataset} & \multirow{2}{*}{\shortstack{\# of nets\\($\times10^3$)}}& \multicolumn{6}{c|}{Approx. \# of nets by degree ($\times10^3$)} & \multirow{2}{*}{\shortstack{Degree of\\largest net}} \\
\cline{3-8}
& & 2-9 & 10-19 & 20-29 & 30-39 & 40-49 & 50 & \\
\hline
\multicolumn{1}{|l|}{Synthetic} & 432 & 63  & 90 & 90 & 90 & 90 & 9 & 50 \\
\hline
\multicolumn{1}{|l|}{OpenROAD} & 104.71 & 104.69 & 0.01 & 0 & 0 & 0 & 0 & 11 \\
\hline
\end{tabular}
\vspace{-0.4cm}
\end{table}

\subsection{Experimental Methodology}

We used the PyTorch library~\cite{pytorch} as the machine learning framework along with the PyTorch Geometric library~\cite{pytorch_geometric}, which is an extension to PyTorch that implements GNN capabilities. One benefit of PyTorch Geometric is that it allows input matrices inside the same batch to be of different sizes. Therefore, it provides a significant runtime improvement due to parallelism across all graph depths.

The final version of our model was trained using the Adam optimizer with a learning rate of \trainLearningRate{}, learning rate reducer with the reduction rate of 0.1 and patience of \trainLRPatience{}, and early stopping with the patience of \trainPatience{} epochs. We picked the best model for evaluation among 100 training runs using the hyperparameters shown in Table~\ref{table:hyperparameters}.


All serial data generation, testing, and evaluation are done on a server with two AMD EPYC 7542 32-core 2.9 GHz processors (128 threads total) and 512GiB DRAM. Serial programs were run on a single core of this machine, and we have not used a graphics processing unit (GPU) or tensor processing unit (TPU). Parallel training, testing, and evaluation are done on an NVIDIA GeForce RTX 4090 24GiB card. 

Other heuristics of \ceff{} modeling are run on our server machine. We compare with the \pimodel{} algorithm of O'Brien et al.~\cite{pi_model} combined the \ceff{} modeling algorithm of Dartu et al~\cite{dmp} as used in OpenROAD. Note, however, that OpenROAD uses pre-characterized tables of \ceff{}, slew, and delay from the Liberty files. Since our synthetic data do not have pre-characterized tables of randomized gate drive strength, we use SPICE simulations to obtain the \ceff{} model's actual slew and delay. This approach makes our analysis of the O'Brien/Dartu heuristic more accurate as it runs SPICE simulations instead of interpolating these tables' values. Although the O'Brien/Dartu heuristic becomes slower this way, we did not consider the execution time of these SPICE simulations in our results. 

The O'Brien/Dartu heuristic can sometimes fail to compute the \ceff{} as it may converge to $C_\text{eff} < 0$ or $C_\text{eff} > C_1+C_2$. In these cases, OpenROAD stops the algorithm and uses lumped capacitance, $C_\text{eff}=C_1+C_2$, for delay calculation. Note that this error cannot occur in GNN-\ceff{} since the last layer of the model has the Sigmoid activation function with a \ceff{}-ratio range of $(0,1]$.

Table~\ref{table:evaluation_dataset_stats} shows both our synthetic and the OpenROAD test datasets. The synthetic dataset provides a uniform distribution of net degrees, while the OpenROAD datasets have real-life examples of nets.


\begin{table*}[!htb]
\def\arraystretch{1.3}
\centering
\caption{Accuracy Results}
\setlength\tabcolsep{4pt}
\label{tab:accuracy_results}
\begin{tabular}{|c|l|r|r|r|r|r|r|r|r|}
\cline{3-10}
\multicolumn{2}{c|}{} & \multicolumn{2}{c|}{Synthetic Test Dataset} & \multicolumn{2}{c|}{\openroadtypical{} Dataset} & \multicolumn{2}{c|}{\openroadfast{} Dataset} & \multicolumn{2}{c|}{\openroadslow{} Dataset} \\ \cline{3-10}

\multicolumn{2}{c|}{} & GNN-\ceff{} & \multicolumn{1}{c|}{O'Brien/Dartu} & GNN-\ceff{} & \multicolumn{1}{c|}{O'Brien/Dartu} & GNN-\ceff{} & \multicolumn{1}{c|}{O'Brien/Dartu} & GNN-\ceff{} & \multicolumn{1}{c|}{O'Brien/Dartu} \\ \hline

& MeAE (fF) & \textbf{\SyntheticAllcasesAbserrorGnnceffMean{}} & \SyntheticAllcasesAbserrorDartuMean{} & \OpenroadTypicalAllcasesAbserrorGnnceffMean{} & \OpenroadTypicalAllcasesAbserrorDartuMean{} & \textbf{\OpenroadFastAllcasesAbserrorGnnceffMean{}} & \OpenroadFastAllcasesAbserrorDartuMean{}  & \textbf{\OpenroadSlowAllcasesAbserrorGnnceffMean{}} & \OpenroadSlowAllcasesAbserrorDartuMean{} \\ \cline{2-10}

& MaAE (fF) & \textbf{\SyntheticAllcasesAbserrorGnnceffMax{}} & \SyntheticAllcasesAbserrorDartuMax{} & \textbf{\OpenroadTypicalAllcasesAbserrorGnnceffMax{}} & \OpenroadTypicalAllcasesAbserrorDartuMax{} & \textbf{\OpenroadFastAllcasesAbserrorGnnceffMax{}} & \OpenroadFastAllcasesAbserrorDartuMax{} & \textbf{\OpenroadSlowAllcasesAbserrorGnnceffMax{}} & \OpenroadSlowAllcasesAbserrorDartuMax{} \\ \cline{2-10}

& MeAER (\%) & \textbf{\SyntheticAllcasesAbserrorratioGnnceffMean{}} & \SyntheticAllcasesAbserrorratioDartuMean{} & \textbf{\OpenroadTypicalAllcasesAbserrorratioGnnceffMean{}} & \OpenroadTypicalAllcasesAbserrorratioDartuMean{} & \textbf{\OpenroadFastAllcasesAbserrorratioGnnceffMean{}} & \OpenroadFastAllcasesAbserrorratioDartuMean{} & \textbf{\OpenroadSlowAllcasesAbserrorratioGnnceffMean{}} & \OpenroadSlowAllcasesAbserrorratioDartuMean{} \\ \cline{2-10}

& MaAER (\%) & \textbf{\SyntheticAllcasesAbserrorratioGnnceffMax{}} & \SyntheticAllcasesAbserrorratioDartuMax{} & \textbf{\OpenroadTypicalAllcasesAbserrorratioGnnceffMax{}} & \OpenroadTypicalAllcasesAbserrorratioDartuMax{} & \textbf{\OpenroadFastAllcasesAbserrorratioGnnceffMax{}} & \OpenroadFastAllcasesAbserrorratioDartuMax{} & \textbf{\OpenroadSlowAllcasesAbserrorratioGnnceffMax{}} & \OpenroadSlowAllcasesAbserrorratioDartuMax{} \\ \cline{2-10}

\rot{\rlap{~~~~~~~All}} & \% of Fails & \textbf{0.00} & \SyntheticFailratio{} & \textbf{0.00} & \OpenroadTypicalFailratio{} & \textbf{0.00} & \OpenroadFastFailratio{} & \textbf{0.00} & \OpenroadSlowFailratio{} \\ \hline

& MeAE (fF) & \textbf{\SyntheticFailAbserrorGnnceffMean{}} & \SyntheticFailAbserrorDartuMean{} & \textbf{\OpenroadTypicalFailAbserrorGnnceffMean{}} & \OpenroadTypicalFailAbserrorDartuMean{} & \textbf{\OpenroadFastFailAbserrorGnnceffMean{}} & \OpenroadFastFailAbserrorDartuMean{} & \textbf{\OpenroadSlowFailAbserrorGnnceffMean{}} & \OpenroadSlowFailAbserrorDartuMean{} \\ \cline{2-10}

& MaAE (fF) & \textbf{\SyntheticFailAbserrorGnnceffMax{}} & \SyntheticFailAbserrorDartuMax{} & \textbf{\OpenroadTypicalFailAbserrorGnnceffMax{}} & \OpenroadTypicalFailAbserrorDartuMax{} & \textbf{\OpenroadFastFailAbserrorGnnceffMax{}} & \OpenroadFastFailAbserrorDartuMax{} & \textbf{\OpenroadSlowFailAbserrorGnnceffMax{}} & \OpenroadSlowFailAbserrorDartuMax{} \\ \cline{2-10}

& MeAER (\%) & \textbf{\SyntheticFailAbserrorratioGnnceffMean{}} & \SyntheticFailAbserrorratioDartuMean{} & \textbf{\OpenroadTypicalFailAbserrorratioGnnceffMean{}} & \OpenroadTypicalFailAbserrorratioDartuMean{} & \textbf{\OpenroadFastFailAbserrorratioGnnceffMean{}} & \OpenroadFastFailAbserrorratioDartuMean{} & \textbf{\OpenroadSlowFailAbserrorratioGnnceffMean{}} & \OpenroadSlowFailAbserrorratioDartuMean{} \\ \cline{2-10}

\rot{\rlap{~~~~Failed}} & MaAER (\%) & \textbf{\SyntheticFailAbserrorratioGnnceffMax{}} & \SyntheticFailAbserrorratioDartuMax{} & \textbf{\OpenroadTypicalFailAbserrorratioGnnceffMax{}} & \OpenroadTypicalFailAbserrorratioDartuMax{} & \textbf{\OpenroadFastFailAbserrorratioGnnceffMax{}} & \OpenroadFastFailAbserrorratioDartuMax{} & \textbf{\OpenroadSlowFailAbserrorratioGnnceffMax{}} & \OpenroadSlowFailAbserrorratioDartuMax{} \\ \hline

& MeAE (fF) & \textbf{\SyntheticNofailAbserrorGnnceffMean{}} & \SyntheticNofailAbserrorDartuMean{} & \OpenroadTypicalNofailAbserrorGnnceffMean{} & \OpenroadTypicalNofailAbserrorDartuMean{} & \OpenroadFastNofailAbserrorGnnceffMean{} & \OpenroadFastNofailAbserrorDartuMean{} & \textbf{\OpenroadSlowNofailAbserrorGnnceffMean{}} & \OpenroadSlowNofailAbserrorDartuMean{} \\ \cline{2-10}

& MaAE (fF) & \textbf{\SyntheticNofailAbserrorGnnceffMax{}} & \SyntheticNofailAbserrorDartuMax{} & \textbf{\OpenroadTypicalNofailAbserrorGnnceffMax{}} & \OpenroadTypicalNofailAbserrorDartuMax{} & \textbf{\OpenroadFastNofailAbserrorGnnceffMax{}} & \OpenroadFastNofailAbserrorDartuMax{} & \textbf{\OpenroadSlowNofailAbserrorGnnceffMax{}} & \OpenroadSlowNofailAbserrorDartuMax{} \\ \cline{2-10}

& MeAER (\%) & \textbf{\SyntheticNofailAbserrorratioGnnceffMean{}} & \SyntheticNofailAbserrorratioDartuMean{} & \textbf{\OpenroadTypicalNofailAbserrorratioGnnceffMean{}} & \OpenroadTypicalNofailAbserrorratioDartuMean{} & \textbf{\OpenroadFastNofailAbserrorratioGnnceffMean{}} & \OpenroadFastNofailAbserrorratioDartuMean{} & \textbf{\OpenroadSlowNofailAbserrorratioGnnceffMean{}} & \OpenroadSlowNofailAbserrorratioDartuMean{} \\ \cline{2-10}

\rot{\rlap{~Non-Failed}} & MaAER (\%) & \textbf{\SyntheticNofailAbserrorratioGnnceffMax{}} & \SyntheticNofailAbserrorratioDartuMax{} & \textbf{\OpenroadTypicalNofailAbserrorratioGnnceffMax{}} & \OpenroadTypicalNofailAbserrorratioDartuMax{} & \textbf{\OpenroadFastNofailAbserrorratioGnnceffMax{}} & \OpenroadFastNofailAbserrorratioDartuMax{} & \textbf{\OpenroadSlowNofailAbserrorratioGnnceffMax{}} & \OpenroadSlowNofailAbserrorratioDartuMax{} \\ \hline

\end{tabular}
\vspace{-0.3cm}
\end{table*}

\subsection{Accuracy Analysis}
\begin{figure}[htb]
    \centering
    \vspace{-0.4cm}
    \includegraphics[width=\linewidth]{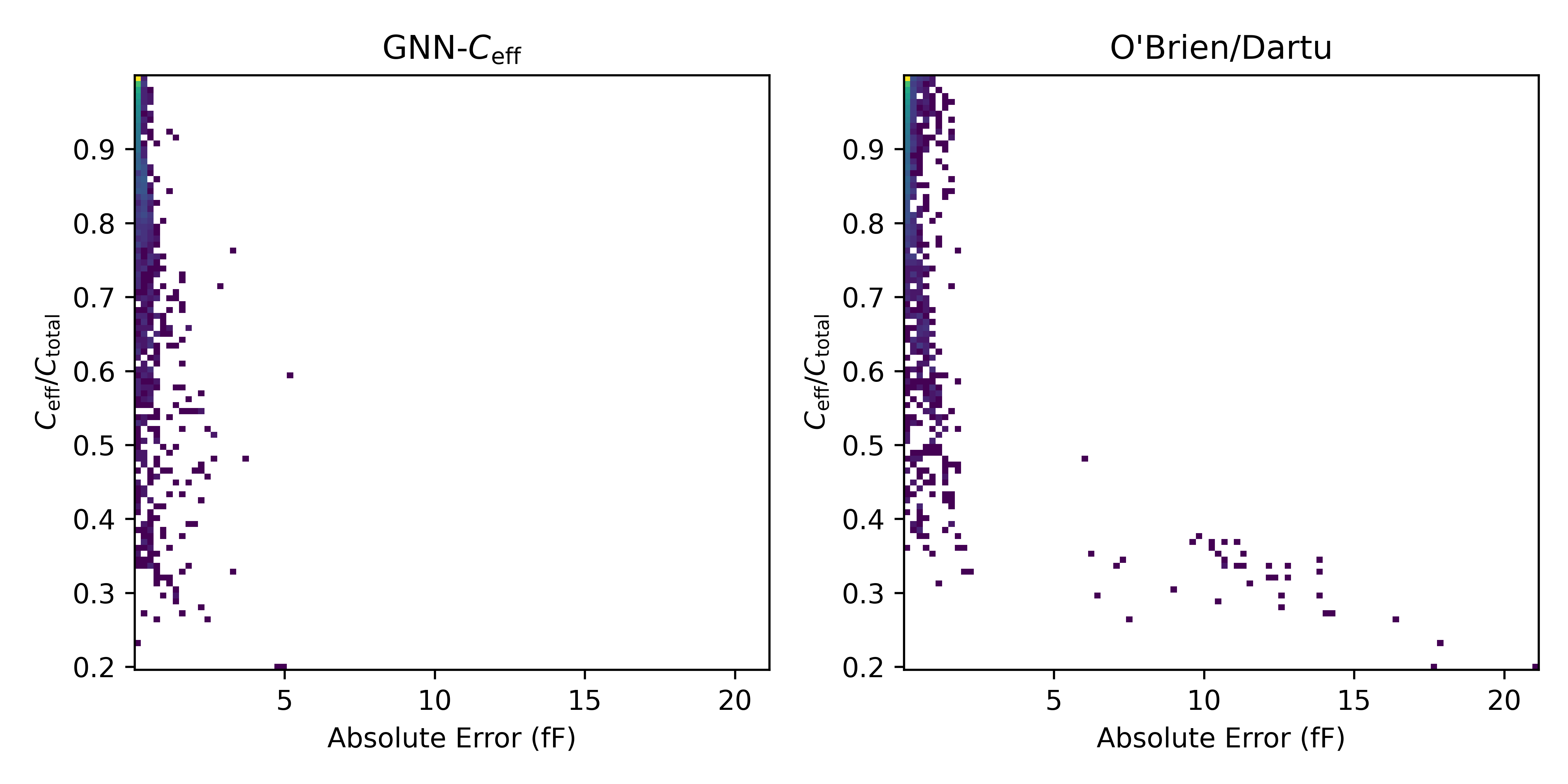}
    \vspace{-0.6cm}
    \caption{The \openroadtypical{} dataset \ceff{} computations of both methods show that the O'Brien/Dartu heuristic performs worse on nets with higher resistive shielding. Note that the 2D histograms' bins are on log scale.}
    \label{fig:resistive_shielding_comparison}
    \vspace{-0.2cm}
\end{figure}

We ran GNN-\ceff{} and the O'Brien/Dartu heuristic on all test datasets and obtained the computed \ceff{}. In Table~\ref{tab:accuracy_results}, mean absolute error (MeAE) and max absolute error (MaAE) are shown in femtofarads. We also added mean absolute error ratio (MeAER) and max absolute error ratio (MaAER), which are $\text{abs}(C_\text{pred} - C_\text{label})/C_\text{label}$. The table is split into three categories: all cases, failed cases, and non-failed cases. The O'Brien/Dartu heuristic can fail to compute \ceff{}, as explained in previous sections. Although GNN-\ceff{} never fails to compute \ceff{}, we still compare the performance of both heuristics on cases where O'Brien/Dartu fails. The synthetic dataset results have larger margins of error than the OpenROAD datasets due to having more nets larger in size. The results show that the O'Brien/Dartu heuristic performs much worse on large nets and has a significantly higher fail rate than the OpenROAD results. This can also be seen in Fig.~\ref{fig:resistive_shielding_comparison}, which shows that the O'Brien/Dartu heuristic has some high errors with very low effective capacitance ratios, yet our method does not.

Another interesting aspect of GNN-\ceff{} is that it does not necessarily need physical wires as features. The model is trained on the synthetic training dataset that is derived from physical wires converted to \rcnetwork{}. The OpenROAD datasets' \gnngraph{}, on the other hand, are created by the parasitic extractor of OpenSTA~\cite{opensta}. This means GNN-\ceff{} can successfully process pure RC networks that also include coupling capacitance.

\subsection{Parallelization Analysis}
Ngspice, GNN-\ceff{}, and the O'Brien/Dartu heuristic are run as previously mentioned and the average execution time results are presented in Table~\ref{tab:time_results}. The datasets are run in the largest batch sizes that can fit in the memory of our GPU. The OpenROAD dataset consists of smaller nets; therefore, we can fit a significantly larger number of problems in a single batch than the synthetic dataset. Execution times are split into three categories: GPU, serial, and parallel. GNN-\ceff{} is run on a GPU, and the other methods are run on a CPU. The serial execution times show the total amount of time it would take to run a batch of problems sequentially. On the other hand, the parallel execution times show the theoretical amount of time if a batch of problems is run on 64 processes in parallel without any overhead like on our large CPU server.

\begin{table}[htb]
\vspace{-0.2cm}
\def\arraystretch{1.5}
\centering
\caption{Runtime Results}
\setlength\tabcolsep{3pt}
\label{tab:time_results}
\begin{tabular}{|c|l|r|r|r|}
\hline

Dataset & \multicolumn{1}{c|}{Method} & \multicolumn{1}{c|}{Batch Size} & \multicolumn{1}{c|}{Execution Time (ms)} & \multicolumn{1}{c|}{Speedup} \\ \hline

\multirow{5}{*}{Synthetic} & \multirow{2}{*}{Ngspice} & 1 & \SyntheticNofailTimeSyntSerialNgspiceMean{} & \SyntheticSpeedupSyntSerialNgspice{} \\ \cline{3-5}
& & 64 & \SyntheticNofailTimeSyntParallelNgspiceMean{} & \SyntheticSpeedupSyntParallelNgspice{} \\ \cline{2-5}
& \multirow{2}{*}{O'Brien/Dartu} & 1 & \SyntheticNofailTimeSyntSerialDartuMean{} & \SyntheticSpeedupSyntSerialDartu{} \\ \cline{3-5}
& & 64 & \SyntheticNofailTimeSyntParallelDartuMean{} & \SyntheticSpeedupSyntParallelDartu{} \\ \cline{2-5}
& GNN-\ceff{} & 6,000 & \textbf{\SyntheticNofailTimeGnnceffMean{}} & - \\ \hline

\multirow{5}{*}{OpenROAD} & \multirow{2}{*}{Ngspice} & 1 & \OpenroadTypicalNofailTimeOpenSerialNgspiceMean{} & \OpenroadTypicalSpeedupOpenSerialNgspice{} \\ \cline{3-5}
& & 64 & \OpenroadTypicalNofailTimeOpenParallelNgspiceMean{} & \OpenroadTypicalSpeedupOpenParallelNgspice{} \\ \cline{2-5}
& \multirow{2}{*}{O'Brien/Dartu} & 1 & \OpenroadTypicalNofailTimeOpenSerialDartuMean{} & \OpenroadTypicalSpeedupOpenSerialDartu{} \\ \cline{3-5}
& & 64 & \OpenroadTypicalNofailTimeOpenParallelDartuMean{} & \OpenroadTypicalSpeedupOpenParallelDartu{} \\ \cline{2-5}
& GNN-\ceff{} & 50,000 & \textbf{\OpenroadTypicalNofailTimeGnnceffMean{}} & - \\ \hline

\end{tabular}
\end{table}

We can immediately notice from the table that SPICE simulations are significantly slower than both GNN-\ceff{} and the O'Brien/Dartu heuristic, even when they are run in parallel. Another interesting finding from the two datasets' execution times is that Ngspice is exponentially slower as the \rcnetwork{} size increases. Although the O'Brien/Dartu heuristic runs faster than Ngspice, it is still slower than GNN-\ceff{}.

GNN-\ceff{} provides \SyntheticSpeedupSyntParallelDartu{} speedup on the synthetic dataset and \OpenroadTypicalSpeedupOpenParallelDartu{} speedup on the OpenROAD dataset over the O'Brien/Dartu heuristic even when the latter is run in parallel with 64 processes.

%% file: conclusion.tex
\section{Conclusion}
\label{sec:conclusion}

In this paper, we proposed GNN-\ceff{}, the first graph neural network to model the effective capacitance of post-layout physical design nets. Due to the nature of GPU parallelization, GNN-\ceff{} can find the effective capacitance of nets in huge batches, providing critical speedup with high accuracy. Our model achieves a mean absolute error of \SyntheticAllcasesAbserrorGnnceffMean{} fF for our synthetic dataset and \OpenroadTypicalAllcasesAbserrorGnnceffMean{} fF for the OpenROAD dataset on the typical corner, while providing a speedup of \SyntheticSpeedupSyntParallelDartu{} for the synthetic dataset and \OpenroadTypicalSpeedupOpenParallelDartu{} for the OpenROAD dataset over the state-of-the-art heuristic run in parallel.